\documentclass[final,journal,twoside]{IEEEtran}
\usepackage{amsmath,amsfonts}
\usepackage{algorithmic}
\usepackage{algorithm}
\usepackage{array}
\usepackage[caption=false,font=normalsize,labelfont=sf,textfont=sf]{subfig}
\usepackage{textcomp}
\usepackage{stfloats}
\usepackage{url}
\usepackage{verbatim}
\usepackage{graphicx}
\usepackage{cite}
\usepackage{stfloats}
\usepackage{makecell}

\usepackage{ragged2e} 
\usepackage{booktabs,makecell, multirow, tabularx}
\usepackage{colortbl}
\usepackage{hyperref}

\newcommand{\eg}{\emph{e.g.},~}
\newcommand{\etal}{\emph{et al}.~}

\usepackage{amsthm}

\usepackage{bbding}
\newtheorem{mypro}{Proposition}
\newtheorem{remark}{Remark}

\begin{document}

\title{Point Cloud Denoising With Fine-Granularity Dynamic Graph Convolutional Networks} 

\author{
Wenqiang~Xu, Wenrui~Dai,~\IEEEmembership{Member,~IEEE,} 
Duoduo~Xue,~\IEEEmembership{Student Member,~IEEE,}
Ziyang~Zheng,~\IEEEmembership{Member,~IEEE,} 
Chenglin~Li,~\IEEEmembership{Member,~IEEE,}
Junni~Zou,~\IEEEmembership{Member,~IEEE,}
and~Hongkai~Xiong,~\IEEEmembership{Senior Member,~IEEE}
\thanks{This work was in part supported by National Natural Science Foundation of China, under Grant 61932022.}
\thanks{Wenqiang~Xu, Wenrui~Dai, and Junni~Zou are with the Department of Computer Science and Engineering, Shanghai Jiao Tong University, Shanghai 200240, China (e-mail: xuwenqiang\_cs@sjtu.edu.cn; daiwenrui@sjtu.edu.cn; zoujunni@sjtu.edu.cn).}
\thanks{Duoduo~Xue, Ziyang~Zheng, Chenglin~Li and Hongkai~Xiong are with the Department of Electronic Engineering, Shanghai Jiao Tong University, Shanghai 200240, China (e-mail: xueduoduo@sjtu.edu.cn; zhengziyang@sjtu.edu.cn; lcl1985@sjtu.edu.cn; xionghongkai@sjtu.edu.cn).}
}


\maketitle

\begin{abstract}
Due to limitations in acquisition equipment, noise perturbations often corrupt 3-D point clouds, hindering down-stream tasks such as surface reconstruction, rendering, and further processing. Existing 3-D point cloud denoising methods typically fail to reliably fit the underlying continuous surface, resulting in a degradation of reconstruction performance. This paper introduces fine-granularity dynamic graph convolutional networks called GD-GCN, a novel approach to denoising in 3-D point clouds. 
The GD-GCN employs micro-step temporal graph convolution (MST-GConv) to perform feature learning in a gradual manner. 
Compared with the conventional GCN, which commonly uses discrete integer-step graph convolution, this modification introduces a more adaptable and nuanced approach to feature learning within graph convolution networks.
It more accurately depicts the process of fitting the point cloud with noise to the underlying surface by and the learning process for MST-GConv acts like a changing system and is managed through a type of neural network known as neural Partial Differential Equations (PDEs). This means it can adapt and improve over time.
GD-GCN approximates the \emph{Riemannian metric}, calculating distances between points along a low-dimensional manifold. This capability allows it to understand the local geometric structure and effectively capture diverse relationships between points from different geometric regions through geometric graph construction based on Riemannian distances. Additionally, GD-GCN incorporates robust graph spectral filters based on the Bernstein polynomial approximation, which modulate eigenvalues for complex and arbitrary spectral responses, providing theoretical guarantees for BIBO stability. Symmetric channel mixing matrices further enhance filter flexibility by enabling channel-level scaling and shifting in the spectral domain.
To our best knowledge, GD-GCN is the first 3-D point cloud denoising method that uses micro-step precision graph convolution in graph representation learning. It addresses the challenge of point cloud denoising from the viewpoint of a dynamic system and achieves the best possible reconstruction of the surface underneath. Experimental results demonstrate that the proposed GD-GCN outperforms state-of-the-art noise removal methods in several ways.
\end{abstract}


\section{Introduction}
\IEEEPARstart{P}{oint} clouds provide a ubiquitous representation of 3-D objects and scenes. The rapid development of 3-D sensing, laser scanning, and render processing techniques has greatly eased the acquisition of 3-D point clouds. However, due to the limitation on the accuracy of acquisition equipment (\eg LiDAR sensors and Kinect), noise and outliers are inevitably introduced in the acquired point clouds and hamper the geometric representation for downstream tasks such as object detection~\cite{9438625,9504551},  part segmentation~\cite{9810920}, and surface reconstruction\cite{9496619,9493165}. There are increasing demands for point cloud denoising to mitigate the corrupted geometric representation caused by noise and outliers. 

Point cloud denoising aims to identify the underlying surface that reflects the real-world geometry of an object from the noisy raw input while preserving the structures like sharp edges and smooth surfaces. Existing attempts~\cite{digne2017bilateral,zhang2019point,alexa2003computing,oztireli2009feature,guennebaud2007algebraic,lipman2007parameterization,huang2013edge,cazals2005estimating,avron2010l1,sun2015denoising,mattei2017point,zeng20193d,dinesh20183d,schoenenberger2015graph,rakotosaona2020pointcleannet,duan20193d,hermosilla2019total,luo2020differentiable,luo2021score} fit the underlying surface with noisy point clouds but are challenged by two main problems: i) approximation of the underlying surface without distorting the data or losing the geometric details, and ii) robustness to noisy point clouds with different noise levels and characteristics.

Optimization-based methods use prior knowledge to preserve geometry details and estimate the underlying surface. Kernel functions are formulated in~\cite{digne2017bilateral,zhang2019point,alexa2003computing,oztireli2009feature,guennebaud2007algebraic,lipman2007parameterization,huang2013edge,cazals2005estimating} to reconstruct the surface more accurately but can only perform well at low noise levels. Sparse priors and ${\ell}_{0}$ or ${\ell}_{1}$ constrained minimization techniques~\cite{avron2010l1,sun2015denoising,mattei2017point} have been applied for denoising, but they struggle with high noise levels due to poor normal estimation on heavily distorted surfaces. Graph signal processing methods~\cite{zeng20193d,dinesh20183d,schoenenberger2015graph} treat each point as a node and connect edges with weights by finding nearest neighbors in Euclidean space. Regularizers based on Graph Total Variation (GTV) and patch-based graph Laplacian have been developed to better preserve sharp features. However, these graph-based methods are computationally intensive and sensitive to parameter settings.


Learning-based methods can be categorized into two kinds. The first kind of methods~\cite{rakotosaona2020pointcleannet,duan20193d,hermosilla2019total} use permutation-invariant functions for blending operations on single points and globally symmetric aggregation operations (e.g., max-pooling) as in PointNet~\cite{qi2017pointnet}. However, these permutation-invariant functions often lack the ability to capture local geometric information. The second kind of methods~\cite{pistilli2020learning,dgcnn} use graph convolution~\cite{bronstein2017geometric}, which is inherently permutation-invariant and compatible with irregular domains. Unlike methods that work on fixed-size patches or apply global operations~\cite{rakotosaona2020pointcleannet,duan20193d,hermosilla2019total}, graph convolution describes the structure of the point cloud as arbitrary neighborhoods for each central point. However, as the depth of the graph convolution layers increases, it can easily lead to signal degradation and over-smoothing.

\begin{figure*}[!t]
\renewcommand{\baselinestretch}{1.0}
\setlength{\abovecaptionskip}{0pt}
\centering
\includegraphics[width=0.9\linewidth]{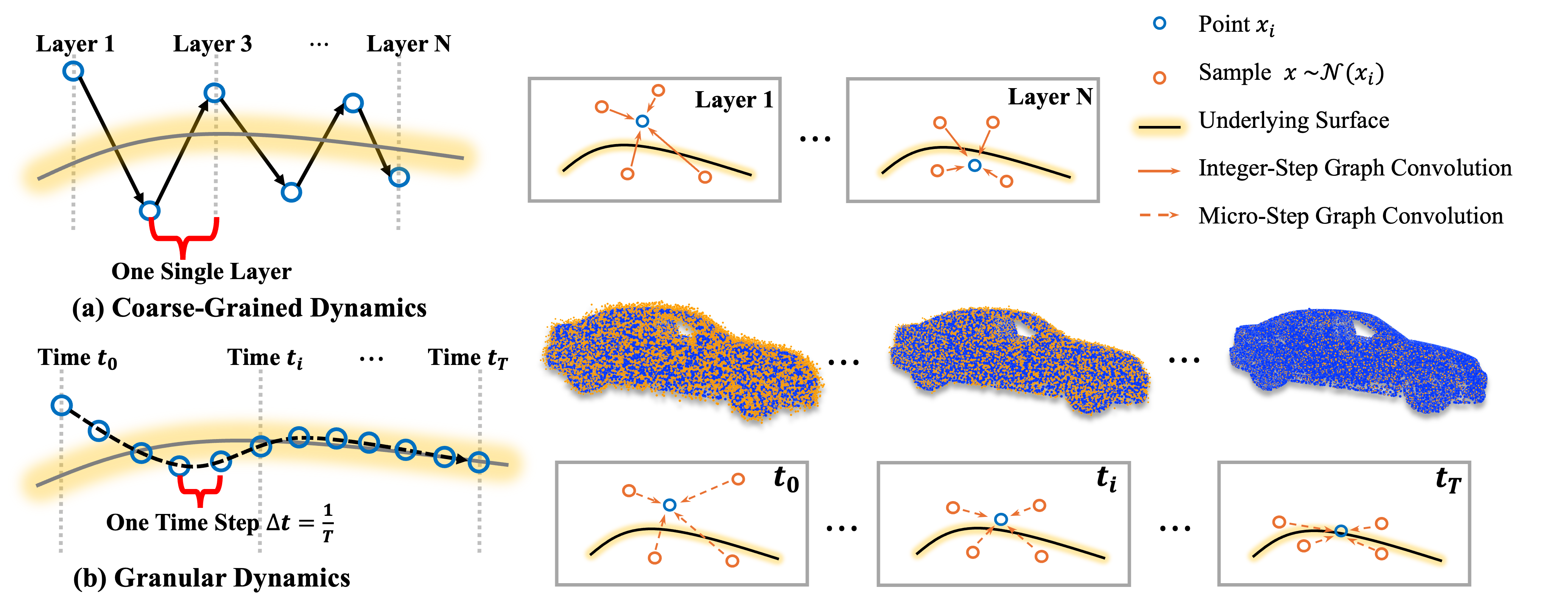}
\caption{(a) Evolving trajectory of the noise points relative to the underlying surface by coarse-grained dynamics GCN. (b) Evolving trajectory of the noise points relative to the underlying surface by fine-granularity dynamic GCN.}\label{pic1}
\end{figure*}

Graph Convolutional Networks (GCN) are commonly used as models that apply discrete integer-step filters on graphs, but they face two main challenges. The first issue is related to the interpretation of GCN as a sequence of one-hop graph filters. This integer-step approach is coarse-grained and insufficient in dynamically capturing the changing trajectory of noisy points. During the denoising process with multi-layer GCN, points tend to “fluctuate up and down” on the underlying surface, as illustrated in the left part of Fig.~\ref{pic1}(a). Therefore, the Multi-layer GCN can cause a point to become increasingly distant from the actual underlying surface when the oscillating points reach some critical points. This will inevitably result in signal degradation and over-smoothing.
The second issue concerns the geometric approach to setting graph edge weights. A common method is to use 3-D Euclidean distance, but this may fail to capture the geometric features of objects, especially at their boundaries. For instance, two points that are close in Euclidean distance may belong to different planes and thus have different geometric properties. On the other hand, using Euclidean distance in a high-dimensional hidden state space may suffer from the curse of dimensionality~\cite{koppen2000curse}, which decreases the effectiveness of the distances in distinguishing between different points.

In this paper, we propose a novel fine-granularity dynamic graph convolutional
networks~(GD-GCN) to address the aforementioned issues. Unlike conventional methods that utilize fixed integer steps for message passing along edges, as illustrated in \figurename~\ref{pic1}(a), our approach decomposes the message passing process into multiple micro-precision steps. This is shown in \figurename~\ref{pic1}(b), where the edge weight (relationship between two points) varies at different moments. We also explore the use of Riemannian distance as a more effective representation of the geometric structure of point cloud data embedded in manifold space and apply robust graph spectral filtering in this process. This micro-step temporal message propagation offers greater flexibility than integer-step graph convolution and more accurately approximates the desired smooth path from noisy points to the underlying surface, as illustrated in the left part of \figurename~\ref{pic1}(b).

To summarize, the contributions of this work include:
\begin{itemize}
    \item We propose the MST-GConv module, which performs temporal graph convolution in fine-grained, micro-steps, enabling a detailed and complete interpretation of noise movements across the underlying graph surface. 
    In comparison to traditional graph convolutional networks (GCN), the GD-GCN can detect topology changes while passing messages in micro precision steps, and the weight of edges is dynamically updated in real-time. The continuity of the dynamics allows for a more precise denoising process that minimizes signal attenuation. The outcomes of this process are solved numerically using neural partial differential equations (neural PDEs).
    \item We propose a geometric graph construction method that learns a \emph{Riemannian metric} and builds a graph on a manifold space. 
    This method captures the geometric features of the point cloud and reflects the true similarity between points better than the Euclidean distance.
    By increasing the distance between points with different geometric properties, our method preserves the underlying structure of the point cloud and enhances the performance of the algorithm.
    \item  We design a robust graph spectral filtering mechanism using innovative polynomial graph spectral filters, featuring complex channel-wise frequency responses optimized for MST-GConv. This approach leverages a Bernstein polynomial basis with trainable coefficients, along with symmetric channel mixing weights, to efficiently scale and shift the spectral components.
    Experimental results demonstrate that our graph spectral filtering method for MST-GConv significantly outperforms traditional spatial graph convolution methods.
\end{itemize}

To our best knowledge, this paper is the first to introduce continuous micro-step temporal graph convolution for 3-D point cloud denoising. By combining geometric graph construction and robust graph spectral filtering, our proposed GD-GCN achieves a significant improvement over the state-of-the-art methods in extensive experiments and demonstrates its capability to remove noise perturbation from point clouds with various statistics.

The rest of the paper is organized as follows. In Section~\ref{sec:related}, we review related work on point cloud denoising. Section~\ref{sec:method} elaborates the proposed method, including the micro-step temporal graph convolution (MST-GConv), geometric graph construction, and robust graph spectral filtering. In Section~\ref{sec:Experiment}, we present experimental results, demonstrating the effectiveness of our approach through quantitative and qualitative analyses. Finally, Section~\ref{sec:con} concludes the paper.

\section{Related Work}\label{sec:related}
\subsection{Optimization-based Denoising}
Optimization-based denoising relies on the geometric priors of objects, including smoothing normal vectors and building graphs that accurately reflect the local geometric shape. They need to avoid over-sharpening or over-smoothing and generalize to diverse inputs. Optimization-based methods can be classified into three categories, \emph{i.e.}, surface fitting, sparse representation, and graph-based methods.

\subsubsection{Surface Fitting} 
Digne \etal\cite{digne2017bilateral}  employed the bilateral filter designed for image denoising \cite{tomasi1998bilateral} in point clouds based on point position and normal. Zhang \etal\cite{zhang2019point} designed a bilateral filter for point cloud denoising by considering the point color.
Alexa \etal\cite{alexa2003computing} pioneered in iteratively projecting noisy points onto natural surfaces and approximated a smooth manifold surface from noisy points using the moving least squares (MLS) method. The projection was extended to preserve the features~\cite{oztireli2009feature,guennebaud2007algebraic}.
Lipman \etal\cite{lipman2007parameterization} designed a locally optimal projection (LOP) operator for parameter-free surface approximation to remove dependence on the estimated surface normal. 
Huang \etal\cite{huang2013edge} developed an edge-aware resampling (EAR) algorithm to generate noise-free point sets with sharp feature preservation, while Cazals \etal\cite{cazals2005estimating} proposed a jets-based local representation using truncated Taylor expansion for accurate differential property estimation on smooth manifolds, especially in low-noise conditions.
However, these methods require the underlying surface to accurately identify point cloud normals  and are highly sensitive to outliers.

\subsubsection{Sparse Representation} Sparse representation is developed under the piecewise constant assumption for common surfaces. Avron \etal\cite{avron2010l1} formulated a global ${\ell}_{1}$-sparse optimization for 3-D surfaces by supposing that the residual of objective function is intensely concentrated near sharp features and the global nature of optimization usually yields a sparse solution. Sun \etal\cite{sun2015denoising} adopted the ${\ell_0}$-norm to induce sparser solution to surface reconstruction that simultaneously recovers sharp features and smoothes the remaining regions than the ${\ell}_{1}$- or ${\ell}_{2}$-norm. 
Mattei \etal\cite{mattei2017point} presented moving robust principal components analysis (MRPCA) to preserve sharp features via a weighted ${\ell}_{1}$-minimization. However, these models are not robust to high noise levels and suffer from over-smoothing or over-sharpening.

\subsubsection{Graph Based Methods} 
Graph signal processing~\cite{shuman2013emerging} has also been introduced into point cloud denoising. 
Zeng \etal\cite{zeng20193d} introduced a discrete patch distance measure and a patch-based graph Laplacian regularizer to approximate manifold dimensions for graph construction. Dinesh \etal\cite{dinesh20183d} developed a reweighted graph Laplacian regularizer (RGLR) for surface normals, offering rotation invariance, enhanced piecewise smoothness, and improved optimization efficiency.
Schoenenberger \etal\cite{schoenenberger2015graph} interpreted the point coordinates as graph signal and employed modern convex optimization methods for filtering and denoising. Nevertheless, these methods highly depend on graph construction. Moreover, spatial graph filtering for iterative approximation of underlying surface is not optimal in reconstruction quality and computation efficiency.

\subsection{Learning-based Denoising} 
Learning-based methods leverage neural networks for feature learning and can be categoried into PointNet-based and convolution-based methods.
\subsubsection{PointNet-based Methods} 
PointNet~\cite{qi2017pointnet} can be directly employed on point clouds to independently learn features for each point with shared multi-layer perceptrons (MLPs). To capture the local information between points, Qi \etal\cite{qi2017pointnetplusplus} proposed a hierarchical network PointNet++ to capture fine geometric structures from the neighborhood of each point. 
Guerrero \etal\cite{guerrero2018pcpnet} introduced PCPNet, a localized PointNet variant that learns \(k\)-dimensional feature vectors for local geometric feature regression, while Hermosilla \etal \cite{hermosilla2019total} proposed the TotalDenoising framework, leveraging a spatial prior for unsupervised 3D point cloud denoising using only noisy data.
Luo \etal \cite{luo2020differentiable, luo2021score} presented an autoencoder-like neural network to learn correction vectors or gradient ascent scores, and Chen \etal \cite{chen2022deep} further introduced regularization priors. Wei \etal \cite{wei2024pathnet} developed a reinforcement learning based path-selective paradigm to dynamically select the most appropriate denoising path for each point and align the selected path with the underlying surface. 
PFN~\cite{9903481} and FCNet~\cite{10099466} integrated filtering techniques with deep learning using the PointNet backbone, where PFN generates learned coefficient vectors to filter noisy points locally, and FCNet employs a non-local filter leveraging self-similarity in point clouds, though its constrained receptive field challenges local smoothing.

\subsubsection{Convolution-based Methods} 
Convolution-based methods enhance the message exchange between individual points. They either project the input point cloud into 2D images \cite{roveri2018pointpronets} to employ convolutional neural networks (CNNs) or represent it as graph signal for filtering with graph convolutional networks (GCN)~\cite{pistilli2020learning,pistilli2020learningrobust}. 
PointProNet~\cite{roveri2018pointpronets} consolidates point collections into a 2D planar embedding, denoises them with image convolutions, and projects the results back to the surface using CNNs, while GPDNet~\cite{pistilli2020learning} uses graph-convolutional layers with single-point convolutions, residual blocks, and dynamic $k$-NN graphs to build feature hierarchies for point cloud denoising in a high-dimensional space.

However, geometric information could be lost due to 3D-to-2D mapping for CNNs, while the signal could collapse rapidly using multi-layer GCN.

\section{Proposed Method}\label{sec:method}
\subsection{Problem Formulation and Motivation}
Consider a noisy point cloud $\mathcal{P}$ containing $N$ points in the 3-D space is acquired by perturbing the underlying manifold $\mathcal{M}$ in the $d$-dimensional manifold space $\mathbb{M}$. 
We leverage an undirected graph \( \mathcal{G} = (\mathcal{V}, \mathcal{E}) \) to represent the point cloud $\mathcal{P}$, where \( \mathcal{V} = \{v_1, \ldots, v_N\} \) is the set of $N$ nodes corresponding to the $N$ points and \(\mathcal{E} = \{e_{ij}, 1 \leq i \neq j \leq N \}\) is the set of edges indicating the connections or relations between points. The topology of $\mathcal{G}$ is encoded by the adjacency matrix $\mathbf{A}_\mathcal{G}\in\mathbb{R}^{N \times N}$ whose element in the $i$-th row and $j$-th column $a_{ij}$ is the weight of edge $e_{ij}\in\mathcal{E}$ that reflects the similarity or proximity between the nodes $v_i$ and $v_j$. For denoising, the graph signal on $\mathcal{G}$ is defined as $\mathbf{P}=[\mathbf{p}_1,\cdots,\mathbf{p}_N]\in\mathbb{R}^{N\times d}$, where $\mathbf{p}_i\in\mathbb{R}^d$ is the perturbed spatial coordinate in $\mathbb{M}$ of the $i$-th point in $\mathcal{P}$ and is taken as the signal living on node $v_i$.  

We consider the widely studied residual graph convolutional network (GCN)~\cite{kipf2016semi}. In GCN, $\mathbf{P}$ is available to each layer. In the $l$-th layer,  Let us denote $\mathbf{Z}^l=[\mathbf{z}_1^l,\cdots,\mathbf{z}_N^l]\in \mathbb{R}^{N \times d_{l}}$ as the matrix of \(d_l\)-dimensional node features extracted from $\mathbf{P}$  where $\mathbf{z}_i^l\in \mathbb{R}^{d_{l}}$ is the feature vector of $v_i$ capturing local geometric patterns or high-level representations relevant to denoising. \(\mathbf{Z}^{l}\) is obtained from \(\mathbf{Z}^{l-1}\) and is fed into the $(l+1)$-th layer. 
\begin{equation}\label{eq:inter-layer}
\mathbf{Z}^{l} = \mathbf{Z}^{l-1} + \delta ((\mathbf{A}_{\mathcal{G}}-\mathbf{I})\mathbf{Z}^{l-1}\boldsymbol{\Theta}^{l}),
\end{equation}
where \(\boldsymbol{\Theta}^l\) denotes the learnable parameter matrix, and \(\delta\) represents the nonlinear activation function in the \(l\)-th layer. Without loss of generality, we define $\mathcal{T}_{\mathcal{G}}^l=  \mathbf{I}+\delta \circ\mathcal{C}_{\mathcal{G}}$ as the operator representing the $l$-th layer and $\mathcal{C}_{\mathcal{G}}^l$ the graph convolution operator. The dynamics of a residual GCN containing $l$ layers is described as
\begin{equation}\label{eq:multi-layer}
\mathbf{Z}^{l+1} = \mathcal{T}_{\mathcal{G}}^{l}\circ\mathcal{T}_{\mathcal{G}}^{l-1} \circ \cdots \circ \mathcal{T}_{\mathcal{G}}^{1}\mathbf{P}. 
\end{equation}

\begin{figure}[!t]
\renewcommand{\baselinestretch}{1.0}
\setlength{\abovecaptionskip}{0pt}
\centering
\includegraphics[width=\linewidth]{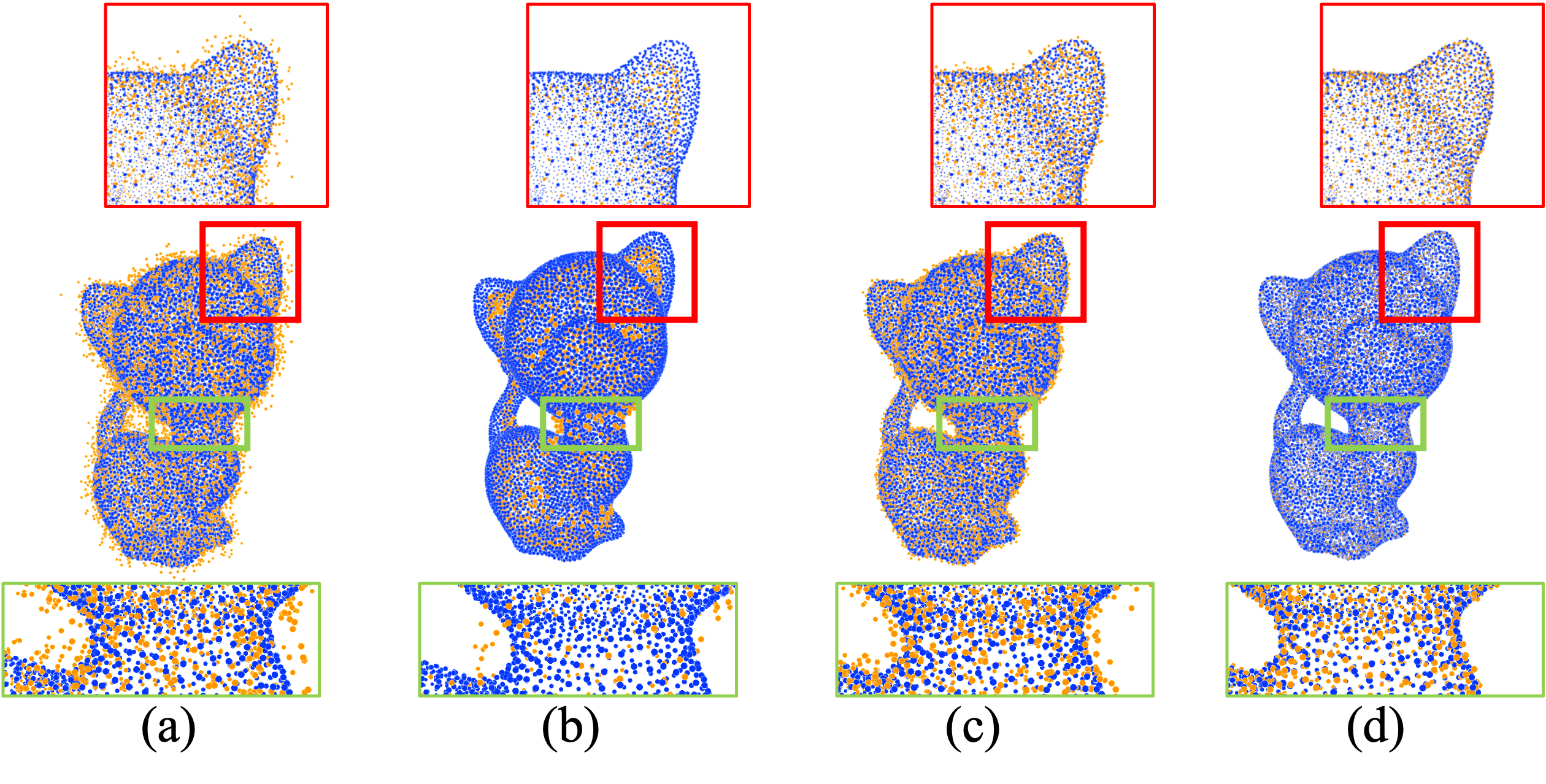}
\caption{
The progression of point cloud denoising. (a) is the noisy input, (b) is the result after integer-step graph convolution, (c) shows the early result of micro-step graph convolution, and (d) shows the final refined output after several micro-steps.
}\label{fig:interpolation}
\end{figure}

According to~\eqref{eq:inter-layer} and~\eqref{eq:multi-layer}, graph convolution for each layer implies one-hop propagation of node features over $\mathcal{G}$. We refer to it as \emph{integer-step graph convolution}. 
When the number of layers increases, GCN that leverages integer-step graph convolution for denoising could suffer from oversmoothing that blends node features together or oversharpening that amplifies noise, due to hop-by-hop message passing over the graph. 
As shown in \figurename~\ref{fig:interpolation}, integer-step graph convolution is more susceptible to noise and outliers in regions with high point density, as closely spaced neighbors may amplify anomalous features. This results in exaggerated deviations in the presence of noise. Conversely, in regions with lower point density, integer-step convolution tends to cause oversmoothing, as the sparse connections between points lead to the premature merging of features, obscuring local structures.

To address these problems, we propose micro-step temporal graph convolution in Section~\ref{sec:MST-GConv}. To enhance the proposed graph convolution, we further develop an approximate Riemannian metric for geometric graph construction in Section~\ref{sec:metric learning} and design robust graph spectral filter in Section~\ref{sec:stable graph spectral filtering}.

\begin{figure*}[!t]
\renewcommand{\baselinestretch}{1.0}
\setlength{\abovecaptionskip}{0pt}
\centering
\includegraphics[width=0.95\linewidth]{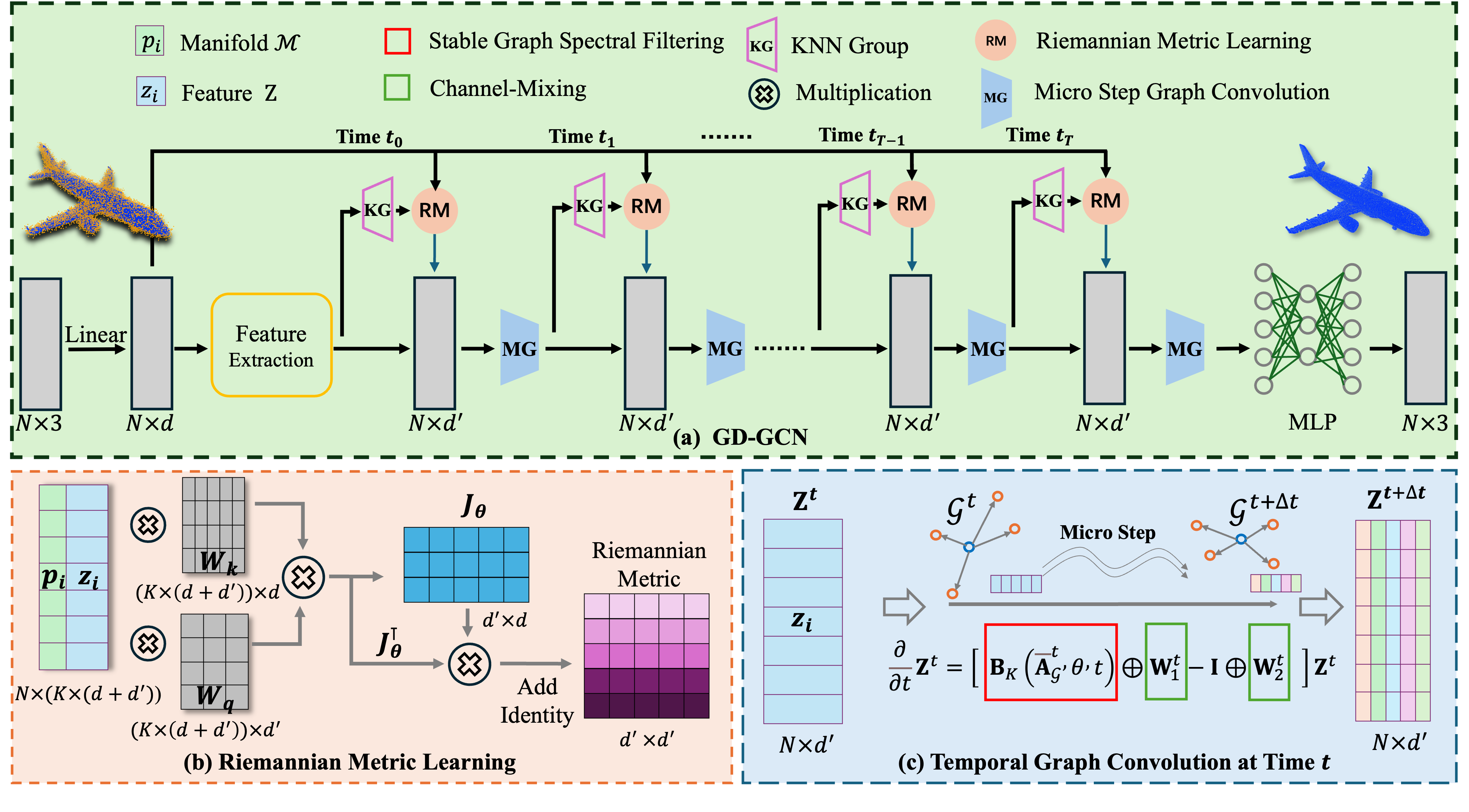}
\caption{(a)~Overall architecture of proposed fine-grained dynamics graph convolution networks (GD-GCN).~(b)~The details of Riemannian metric learning.~(c)~The details of micro-step temporal graph convolution at time \(t\).}\label{architecture}
\end{figure*}

\subsection{Architecture Overview}\label{sec:overview}
We begin with an overview of the proposed geometric fine-grained dynamic graph convolution networks~(GD-GCN). \figurename~\ref{architecture} illustrates the architecture of the proposed GD-GCN. Spatial coordinates $\mathbf{P}$ in the $d$-dimensional manifold space are first obtained from the point cloud using EdgeConv~\cite{wang2019dynamic} and are fed into GD-GCN. For the $l$-th layer, the graph adjacency matrix $\mathbf{A}_\mathcal{G}$ is constructed from $\mathbf{P}$ and $\alpha\mathbf{Z}^l$ with a scale factor $\alpha$.

\figurename~\ref{architecture}(b) elaborates the approximate Riemannian metric for geometric graph construction of MST-GConv to enable fine-grained feature learning on the manifold \(\mathcal{M} \in \mathbb{R}^{d}\) and allow for the construction of deep networks that perform well with large numbers of micro steps. This learned \emph{Riemannian metric} facilitates pulling back the Euclidean distance structure embedded in the manifold \(\mathcal{M}\), enhancing the separation between points with different geometric properties. The geometric graph, based on Riemannian distance, is updated with each micro-step of node updates. Finally, \figurename~\ref{architecture}(c) illustrates the proposed Bernstein polynomial approximation for graph spectral filtering at each micro-step, which increases the flexibility of MST-GConv. The constraints on the eigenvalues' bounds ensure stability, while symmetric channel mixing matrices further differentiate feature channels and effectively address the issue of low-frequency dominance.

\subsection{Micro-Step Temporal Graph Convolution}\label{sec:MST-GConv}

Contrary to one-hop message passing between nodes in~\eqref{eq:inter-layer}, we propose micro-step temporal graph convolution (MST-GConv) that achieves fine-grained temporal message passing between neighboring points in multiple micro steps. We first formulate MST-GConv and then relate it to neural ordinary differential equations (ODEs) for efficient realization.

\subsubsection{Formulation of MST-GConv} 
Without loss of generality, we consider any $l$-th micro step. The node feature $\mathbf{z}_i^l$ for $v_i$ is updated using a micro step size $\tau^l\ll 1$.
\begin{equation}\label{fractional step GCN}
\frac{\mathbf{z}_{i}^{l} - \mathbf{z}_{i}^{l-1}}{\tau^{l}} = \delta ( \sum_{j\in\mathcal{N}^l(i)} a_{ij}^l\left(\mathbf{z}_{j}^{l-1}-\mathbf{z}_{i}^{l-1}\right)\boldsymbol{\Theta}^l).
\end{equation}
Here, $\mathcal{N}^l(i)$ denotes the neighborhood of $v_i$ and $a_{ij}^l$ is the element in the $i$-th row and $j$-th column of the adjacency matrix $\mathbf{A}_{\mathcal{G}}^l$ in the $l$-th micro-step. Given the center point $v_i$, $\mathcal{N}^l(i)$ could vary with $l$ to adapt to temporal evolution of geometric structures at $v_i$. For $j\in\mathcal{N}^l(i)$,  $a_{ij}^l$ is determined with the node features $\mathbf{z}_i^l$ and $\mathbf{z}_j^l$ to represent the similarity between $v_i$ and $v_j$ in the $l$-th micro layer. From~\eqref{fractional step GCN}, the messages propagated for $L$ micro steps over $\mathcal{G}$ can be represented as
\begin{equation}\label{FPTV GCN}
\mathbf{Z}^{L}=\mathbf{Z}^{0}+\sum_{l=1}^{L} \delta( (\mathbf{A}^{l}_{\mathcal{G}}-\mathbf{I})\mathbf{Z}^{l-1}\boldsymbol{\Theta}^{l})\tau^{l}.
\end{equation}

Eq.~\eqref{FPTV GCN} implies that the micro-step size \(\tau\) corresponds to interpolating \(L/\tau\) times along the gradient direction during the graph convolution, with real-time updates ensuring the gradient accurately follows the underlying manifold. As shown in \figurename~\ref{fig:interpolation}, compared to a single integer-step graph convolution, the micro-step graph convolution performs \(1/\tau\) interpolations, generating \(1/\tau\) new adjacency matrices that represent the updated geometric topology. This approach offers a more refined perception of the changes in geometric features within the point cloud compared to integer-step graph convolution.

\subsubsection{Realization of MST-GConv} 
We relate MST-GConv in~\eqref{FPTV GCN} to neural ODEs~\cite{chen2018neuralode,dupont2019augmented,poli2019graph} to allow efficient gradient computation during backpropagation and achieve numerical solutions without storing intermediate states. MST-GConv can be viewed as a parametric diffusion equation discretized by the step size $\tau^l$. It aligns with neural ODEs by introducing the continuous (time) variable $t$ for $l$ and making the step size $\tau^t\rightarrow 0$ in~\eqref{FPTV GCN}.
Let us define \(\mathcal{C}_{\mathcal{G}}^l[\mathbf{A}^{l}_{\mathcal{G}}, \boldsymbol{\Theta}^{l}] = \delta((\mathbf{A}^{l}_{\mathcal{G}} - \mathbf{I})\mathbf{Z}^{l-1} \boldsymbol{\Theta}^{l})\) as the graph convolution operation for the \(l\)-th micro layer. The message passing for a time interval \(\Delta t\) using MST-GConv is described as 
\begin{equation}\label{time window GCN}
\mathbf{Z}^{t_0+\Delta t} = \mathbf{Z}^{t_0} + \int_{t_0}^{t_0+\Delta t} \mathcal{C}_{\mathcal{G}}^t[\mathbf{A}^{t}_{\mathcal{G}}, \boldsymbol{\Theta}^{t}](\mathbf{Z}^t) \, \mathrm{d}{t}.
\end{equation}

We can efficiently solve~\eqref{time window GCN} using discretization schemes such as adaptive-step methods (e.g., \emph{dopri5}) or fixed-step methods (e.g., \emph{rk4}). In this paper, we employ the fixed-step method \emph{rk4} with a constant time step of \(\Delta t = 0.1\), and the activation function is set to the default parameters of PyTorch’s \emph{LeakyReLU}.

Therefore, the proposed MST-GConv framework leverages temporal graph convolution 
\(\mathcal{C}_{\mathcal{G}}^t[\mathbf{A}^{t}_{\mathcal{G}}, \boldsymbol{\Theta}^{t}]\) 
to model autoregressive sequences of time-varying graphs \(\left\{\mathcal{G}_{t}\right\}\), establishing a direct connection with dynamical network theory~\cite{weinan2017proposal,lu2018beyond}. Subsequently, we the framework also updates \(\mathbf{A}^{t}_{\mathcal{G}}\) in real-time to adapt to dynamic changes in geometric topology, as described in the geometric graph construction in Section~\ref{sec:metric learning}. Furthermore, MST-GConv incorporates more expressive graph convolutional filters, enhancing its ability to capture complex structures, as detailed in Section~\ref{sec:stable graph spectral filtering}.

\subsection{Geometric Graph Construction}\label{sec:metric learning}

Graph construction is essential for capturing the topological structure of manifolds by measuring the distance between point features. The popular \(k\)NN algorithm uses the Euclidean distance but overlooks the geometric property of underlying manifold. The Mahalanobis distance struggles with complex manifolds, while the geodesic distance is computation intensive for large-scale or dynamic point clouds. To address this problem, we propose geometric graph construction that leverages an approximate Riemannian distance to better capture the intrinsic geometry of underlying manifold.
\subsubsection{Riemannian Metric for Graph Construction} 
Suppose that 3D point clouds reside on a \( d \)-dimensional manifold \( (\mathcal{M}, g) \) embedded in a higher-dimensional ambient space, where the metric tensor \( g \) defines the intrinsic geometry of the manifold $\mathcal{M}$ to determine the distance on $\mathcal{M}$. We construct a graph $\mathcal{G}$ to discretize $\mathcal{M}$ and utilize message passing along the edges $\xi = \{a_{ij}\}$ to propagate the embedding $\mathbf{X}=(\mathbf{P}, \alpha \mathbf{Z})$ with the scaling factor $\alpha \geq 0$. The graph structure allows for the aggregation of both positional and feature information. We adhere to the Riemannian manifold hypothesis~\cite{lee2018introduction} and leverage the intrinsic Riemannian distance $\ell$ between points on $\mathcal{M}$ to construct the graph. Given a center point $\mathbf{p}_{i}$ and its neighboring point $\mathbf{p}_{j}$, $\ell$ is
calculated as 
\begin{equation}\label{riemannian distance}
\ell= [(\mathbf{p}_{i}-\mathbf{p}_{j})\mathbf{G} (\mathbf{p}_{i}-\mathbf{p}_{j})]^{\frac{1}{2}},
\end{equation}
where $\mathbf{G} \in \mathbb{R}^{d \times d}$ is the \emph{Riemannian metric}~\cite{nomizu1961existence} defined as
\begin{equation}\label{riemannian metric}
\mathbf{G} = \mathbf{I} + \alpha ^{2} (\nabla \mathbf{z}(\mathbf{p}) )^{\top}(\nabla \mathbf{z}(\mathbf{p})).
\end{equation}
In~\eqref{riemannian metric}, the Jacobian matrix \(\nabla \mathbf{z}(\mathbf{p})\) represents the differential of the map \( \mathbf{z}: \mathbb{R}^{d^l} \to \mathbb{R}^d \), which encodes the local transformation of the manifold. 
While neural networks can compute Jacobian-vector products via backpropagation, they are insufficient for capturing the manifold’s intricate local geometry in real time. In contrast, the full Jacobian matrix provides the necessary detail for accurate graph construction. However, this level of detail cannot be explicitly obtained through the current backpropagation mechanisms in deep networks.
To address these limitations, we propose an attention-based mechanism to approximate the full Jacobian matrix.

\subsubsection{Approximation of Riemannian Metric} 
We develop an approximate \emph{Riemannian metric} for geometric graph construction. Given any point $i$, we first leverage a linear transformation (point-wise MLPs) to transform the point coordinates in the Euclidean space to $d$-dimensional coordinates $\mathbf{p}_i$ on the manifold $\mathcal{M} \in \mathbb{R}^{N\times d}$, and then incorporate $\mathbf{p}_i$ and $\mathbf{z}_i^l$ to produce the embedding $\mathbf{x}_{i}^l$ in the $l$-th layer. $\mathbf{X}\in\mathbb{R}^{N\times(d+d^{l})}$ is fed into a grouping layer $g(\cdot)$ \cite{qi2017pointnetplusplus} to identify a fixed number of $K$ neighboring points in the neighborhood for each point. The embeddings of all the $N$ points are collected as $\tilde{\mathbf{X}}\in\mathbb{R}^{N\times(K(d+d^{l}))}$ and are fed into a multi-head attention~\cite{vaswani2017attention}.
The independent outputs are then concatenated and linearly transformed to the expected dimension $\mathbb{R}^{N\times d}$. The learned Jacobian $\mathbf{J}_{\theta}\approx \nabla \mathbf{z}(\mathbf{p})$ is used to substitute $\nabla\mathbf{z}(\mathbf{p})$ in~\eqref{riemannian metric} for calculating $\mathbf{G}$.
\begin{equation}\label{learned jacobian}
\mathbf{J}_{\theta}=\operatorname{LeakyRelu}\left(\frac{ \tilde{\mathbf{X}}\mathbf{W}_{Q} (\tilde{\mathbf{X}}\mathbf{W}_{K})^T } {\sqrt{d}}\right),
\end{equation} 
where \(\mathbf{W}_{K}\) and \(\mathbf{W}_{Q}\) are matrices of learnable parameters. 

Given a point $\mathbf{x}_{i}$ and an arbitrary point $\mathbf{x}_{j}$ in its neighborhood $j\in\mathcal{N}(i)$, the weight of their edge $\tilde{a}_{i j}=\exp{(-|\ell_{ij}|^{2}/(2\delta^{2}))}$ using their Riemannian distance $\ell_{ij}$ in~\eqref{riemannian distance}, where $\delta$ denotes the standard deviation of the Riemannian distance of each point to its neighbors. This method expresses correlations that hold greater significance in preserving geometric features compared to the commonly used Euclidean distance method. It highlights the disparities in geometric features between certain adjacent points such as those located on opposite sides of a sharp angle, while maintaining strong correlations among neighboring points in smooth regions.

\subsection{Spectrum-Adaptive Stable Graph Spectral Filtering}\label{sec:stable graph spectral filtering}

Observing the formulation in Eq.~\eqref{time window GCN}, if we disregard the dynamic adjustments to the adjacency matrix caused by the evolving topology and the effects of the non-linear activation functions, the result simplifies to a diffusion equation:
\[
\mathbf{x}(t) = e^{\overline{\mathbf{A}} t} \mathbf{x}(0),
\]  
where \(\overline{\mathbf{A}}\) represents a general graph filter. It is evident that as time progresses, the eigenvalues associated with MST-GConv grow exponentially, which may lead to numerical instability and loss of control over the spectral behavior. Even when the adjacency matrix is dynamically adjusted in response to the evolving topology, this issue may not be fully resolved, as the exponential growth of eigenvalues can still lead to instability and undesirable spectral distortions. 

To address this, we propose a graph spectral filtering framework with bounded-input bounded-output (BIBO) stability and spectral adaptivity, denoted as \(\mathcal{C}_{\mathcal{G}}^t[\mathbf{A}_{\mathcal{G}}^t, \Theta^t]\). The robust graph filter in MST-GConv is founded on two key principles: stability and adaptivity.  

i) \textbf{Stability}: The graph spectral range is rigorously defined to ensure BIBO stability~\cite{RM291-297}, enabling robust filtering in dynamic systems. Unlike manually designed filters~\cite{gasteiger2018predict, zhu2021interpreting}, which are limited to fixed low-pass or high-pass characteristics without adaptive spectral learning, and learning-based methods~\cite{xu2018representation, liu2020towards, defferrard2016convolutional, bianchi2021graph, he2021bernnet}, which offer limited control over the spectral range, our method provides enhanced flexibility in spectral learning while maintaining bounded outputs. This ensures the stability of dynamic systems.  

ii) \textbf{Adaptivity}: The proposed graph filter dynamically adjusts the frequency components, effectively mitigating the low-frequency dominance (LFD) issue in MST-GConv. LFD often suppresses high-frequency components, leading to structural contraction over time as propagation increases (\(t \to \infty\))~\cite{cai2020note}. By enabling adaptive frequency control, our method achieves a balanced spectral response, preserving structural integrity and ensuring robust performance across varying propagation times.

Bounded-input, bounded-output (BIBO) stability is a fundamental property of systems that process external inputs. In a BIBO-stable system, the output remains bounded for any bounded input. In point cloud denoising, the lack of BIBO stability in a filter—even when applied with micro-step propagation—can lead to noise amplification during propagation, resulting in spurious feature generation. BIBO stability ensures that the output stays within a defined range, preventing excessive smoothing or numerical divergence during filtering. This property is particularly critical for large-scale point cloud data, where stability is essential to preserving the integrity of the original geometric structure.

\begin{table*}[!t]
\renewcommand{\baselinestretch}{1.0}
\renewcommand{\arraystretch}{1.0}
\setlength{\tabcolsep}{0pt}
\setlength{\abovecaptionskip}{0pt}
\centering
\caption{Summary of manually designed and learned models for graph spectral filtering.}\label{different gsp}
\begin{tabular}{@{}cl|c|c|c@{}}
\toprule
\multicolumn{2}{c|}{Graph Spectral Filters} & Spectrum Modulation  & Filter Properties & Spectrum Range \\ 
\midrule
\multirow{4}{*}{Manual} 
& Personalized PageRank~\cite{gasteiger2018predict} & $g(\lambda)=\theta(1-(1-\theta) \lambda)^{-1}, \theta \in(0,1]$  & low-pass & \((\theta, 1] \)  \\ \cline{2-5}  
& GNN-LF~\cite{zhu2021interpreting}  & $g(\lambda)=\frac{(\mu +(1-\theta_1) \lambda}{((\theta_1+{1}/{\theta_2}-1) +(2-\theta_1- {1}/{\theta_2}) \lambda)^{-1}},\theta_1 \in [\frac{1}{2},1), \theta_2\in (0,\frac{2}{3})$ & low-pass  &  \(\left( \frac{\mu}{\theta_1 + {1}/{\theta_2} - 1}, \, \mu - \theta_1 + 1 \right]
\) \\ \cline{2-5} 
& GNN-HF~\cite{zhu2021interpreting} &  $g(\lambda)=\frac{\left(1+\theta_{1}(1-\lambda)\right)}{\left(\left(\theta_{1}+1 / \theta_{2}\right)+\left(1-\theta_{1}-1 / \theta_{2}\right) \lambda\right)^{-1}}, \theta_{1} \in(0, \infty), \theta_{2} \in(0,1]$
& high-pass & $\left[\theta_1\!+\!\frac{1}{\theta_2},(1+\theta_{1})^{2}\!+\!\theta_1\!+\!\frac{1}{\theta_2}\right)$\\ 
\midrule
\multirow{6}{*}{Learned}  
& Vanilla basis~\cite{xu2018representation,liu2020towards}  & $g(\lambda)=\sum_{k=1}^{K} \theta_{k} \lambda^{k}$   & low-pass or high-pass & --   \\ \cline{2-5}  
& \multirow{2}*{Chebyshev basis~\cite{defferrard2016convolutional}} & $g(\lambda)=\sum_{k=0}^{K} \theta_{k} T_{k}(\lambda)$, & \multirow{2}*{arbitrary} & \multirow{2}*{--} \\ 
&&$T_{k}(\lambda)=2 \lambda T_{k-1}(\lambda)-T_{k-2}(\lambda),T_{0}=1, T_{1}=\lambda$ & &\\
\cline{2-5} 
& ARMA basis~\cite{bianchi2021graph} & $g(\lambda)=\frac{\sum_{k=0}^{K-1} \theta_1 \lambda^{k}}{1+\sum_{k=1}^{K} \theta_2\lambda^{k}}$  & arbitrary & --  \\ \cline{2-5}  
& Bernstein basis~\cite{he2021bernnet}       & $g(\lambda)=\sum_{k=0}^{K} \theta_k \cdot \binom{K}{k}(1-\lambda)^{K-k} \lambda^{k}, \theta_{k}>0$  & arbitrary & $\left(0,  \sum \theta_k2^{-K} \binom{K}{\lfloor K/2 \rfloor}\right]$ \\ 
\bottomrule
\end{tabular}
\end{table*}

\subsubsection{Bernstein Polynomial Approximation With BIBO Stability} 
In Table~\ref{different gsp}, we analyze existing handcrafted and learned graph spectral filters. Filters manually designed using personalized PageRank and are restricted to low-pass or high-pass filtering and cannot adapt to arbitrarily varying frequency components. For learned filters, Chebyshev polynomials and ARMA are flexible in filtering characteristics but cannot constrain the range of spectrum, while Bernstein polynomials have the advantages in constraining the range of spectrum to $[0,1)$. Thus, we achieve graph spectral filtering via Bernstein polynomial approximation to obtain a steady-state solution~\cite{chamberlain2021grand} to~\eqref{time window GCN} and ensure the numerical stability in each micro step. 
From~\eqref{time window GCN}, at time $t$, we achieve the convolution $\mathcal{C}_{\mathcal{G}}^t$ for obtaining $\mathbf{Z}^{t+\Delta t}$ with  
To avoid complex eigen-decomposition \({\mathbf{A}}^{t}_{\mathcal{G}}=\mathbf{U} \boldsymbol{\Lambda} \mathbf{U}^{T}\), we approximate the kernel function \(g(\lambda)\) using a \(K\)-order Bernstein polynomial \(\mathbf{B_{K}}(\lambda)\)~\cite{he2021bernnet} constructed by the Bernstein polynomial basis $\{b_{k}^{K}(\lambda)\}_{k=0}^{K}$~\cite{farouki2012bernstein} with learnable coefficients $\{\theta_{k}\}_{k=0}^{K}$:
\begin{equation}\label{bernstein polynomial}
\mathbf{B_{K}}(\lambda)\!=\!\sum_{k=0}^{K}\!\theta_{k}b_{k}^{K}(\lambda)\!=\!\sum_{k=0}^{K}\!g\!\left(\frac{k}{K}\right)\!\binom{K}{k}(1\!-\!\lambda)^{K\!-\!k}\lambda^{k}.
\end{equation} 
When \(\theta_{k} = g(k/K)\), \(\mathbf{B_{K}}(\lambda)\) converges to \(g(\lambda)\) as \(K \to \infty\)~\cite{farouki2012bernstein}. 
We impose constraints on the coefficients to ensure \(\mathbf{B_{K}}(\lambda)\) maintains computable bounds without losing spectral response flexibility, ensuring the stability of MST-GConv.

\begin{mypro}\label{pro:1}
Let \(\mathbf{B_{K}}(\lambda)=\sum_{k=0}^{K} \theta_{k}\binom{K}{k}(1-\lambda)^{K-k} \lambda^{k}\) be a Bernstein polynomial on \(\lambda \in (0,1]\). Given non-negative graph filter $g$, when $\theta_{k} = g(k/K)2^{-K}(K+1)!((K-\lfloor K/2\rfloor)!)^{-1}$ for any $k\in\mathbb{N}$, $0 < \mathbf{B_{K}}(\lambda) \leq 1$.
\begin{IEEEproof}
Please refer to Appendix~\ref{app:1}.
\end{IEEEproof}
\end{mypro}

\subsubsection{Spectrum Adaptation With Channel Mixing} 
Although BIBO-stable spectral filtering ensures system stability without numerical divergence, repeated filtering within a fixed bounded spectral domain inevitably leads to the LFD phenomenon as the network depth increases~\cite{diunderstanding}. We incorporate learnable channel mixing matrices to adaptively adjust the frequency components in the graph spectral domain and mitigate the LFD phenomenon. The time derivative is formulated as
\begin{equation}\label{MST-GConv—W}
\frac{\partial}{\partial t} \mathbf{Z}^{t}= \mathbf{B_{K}}\left(\mathbf{A}^{t}_{\mathcal{G}}, \theta, t\right) \mathbf{Z}^{t} \mathbf{W}_1^{t} - \mathbf{Z}^{t}\mathbf{W}_2^{t}.
\end{equation} 

\begin{remark}
Using symmetric \(d \times d\) channel mixing matrices \(\mathbf{W}_1^t\) and \(\mathbf{W}_2^t\) enables scaling and shifting of the graph spectral filtering process, allowing the model to control the spectral response across channels and effectively address LFD by emphasizing specific frequency components.
\end{remark}

Let $\boldsymbol{\phi} \in \mathbb{R}^{N}$ denote the vector of eigenvalues of the graph filter \(\mathbf{B_{K}}\), and $\boldsymbol{\mu} \in \mathbb{R}^{d}$ and $\boldsymbol{\varphi} \in \mathbb{R}^{d}$ the vectors of eigenvalues of channel mixing matrices $\mathbf{W}_1^{t}$ and $\mathbf{W}_2^{t}$, respectively. By vectorizing \(\mathbf{Z}^{t}\) as \(\mathrm{vec}(\mathbf{Z}^{t}) \in \mathbb{R}^{N \times d}\) and utilizing the properties of the Kronecker product and sum, the spectrum modulation of the graph filter can be expressed as:
\begin{equation}
\texttt{Eigen}\left((\mathbf{W}_1^{t})^\top \otimes \mathbf{B_{K}} \oplus (-\mathbf{W}_2^{t})^\top \right) = \boldsymbol{\mu} \otimes \boldsymbol{\phi} - \boldsymbol{\varphi} \otimes \boldsymbol{1},
\end{equation}
where \(\boldsymbol{1} \in \mathbb{R}^{N}\) represents a vector of ones. The adjusted eigenvalues, expressed in \(\mathbb{R}^{N \times d}\), are obtained from the outer product of the eigenvalues of the graph filter and the channel mixing matrices. For each channel feature, spectral adjustments—comprising both scaling and translation—are applied independently. This method introduces negative frequency components while preserving the BIBO stability of graph spectral filtering. It prevents graph convolution from converging within a fixed positive spectral range to enhance specific spectral components and reduce low-frequency dominance.
\subsection{Loss Functions}
We propose loss functions for end-to-end training of GD-GCN in both supervised and unsupervised settings. 
We denote the input degraded point cloud as \( \mathcal{U} = \{ \mathbf{u}_i \}_{i=1}^N \) and the ground truth point cloud as \( \hat{\mathcal{U}} = \{ \hat{\mathbf{u}}_{i} \}_{i=1}^N \).

\subsubsection{Supervised Training}
The supervised training loss $\mathcal{L}_{\rm SL}$ combines Chamfer distance (CD) $\mathcal{L}_{\rm CD}$~\cite{fan2017point} that calculates the average distance of denoised points from the clean surface and Earth Mover’s distance (EMD) $\mathcal{L}_{\rm EMD}$~\cite{fan2017point} that measures the minimum distance between denoised points $\hat{\mathcal{U}}$ and clean points $\mathcal{U}$ for all bijections $\varphi: \hat{\mathcal{U}} \rightarrow \mathcal{U}$.
\begin{align}
&\mathcal{L}_{\rm SL}\!=\!\mathcal{L}_{\rm CD} + \mathcal{L}_{\rm EMD},\\
&\mathcal{L}_{\rm CD}\!=\!\frac{1}{|\hat{\mathcal{U}}|}\!\sum_{\hat{\mathbf{u}}\in\hat{\mathcal{U}}}\!\min_{i}\!\|\hat{\mathbf{u}}_{i}\!-\!\mathbf{u}_{j}\|^{2\!}\!+\!\frac{1}{|\mathcal{U}|}\!\sum_{\mathbf{u}\in \mathcal{U}}\!\min_{j}\!\|\hat{\mathbf{u}}_{i}\!-\!\mathbf{u}_{j}\|^{2},\label{CD distance}\\
&\mathcal{L}_{\rm EMD}\!=\!\min_{\varphi: \hat{\mathcal{U}} \rightarrow \mathcal{U}} \frac{1}{|\hat{\mathcal{U}}|} \sum_{\mathbf{u} \in \hat{\mathbf{u}}}\|\hat{\mathbf{u}}-\varphi(\mathbf{u})\|^{2}.\label{EMD distance}
\end{align}

\subsubsection{Unsupervised Training}
The unsupervised training loss $\mathcal{L}_{\rm UL}$ combines the repulsion loss $\mathcal{L}_{\rm repul}$ that prevents clumping and maintains local structure and reconstruction loss $\mathcal{L}_{\rm recons}$ that aligns denoised points with the underlying surface.
\begin{align}
&\mathcal{L}_{\rm UL}= \mathcal{L}_{\rm recons} + \lambda \cdot \mathcal{L}_{\rm repul},\\
&\mathcal{L}_{\rm recons}= \frac{1}{|\mathcal{U}|} \sum_{\mathbf{u} \in \mathcal{U}} \mathbb{E}_{\mathbf{v} \sim \mathbb{P}(\mathbf{v} \mid \mathbf{u})} \left\| \mathbf{u} - \mathbf{v}) \right\|^{2},\\
&\mathcal{L}_{\rm repul}= \frac{1}{|\mathcal{U}|}  \sum_{\mathbf{u} \in \mathcal{U}} \sum_{j \in \mathcal{N}(i)} r_{i,j} \left\| \mathbf{u}_{i} - \mathbf{u}_{j} \right\|^{2}.
\end{align}
Here, $\lambda$ is set to 0.01, $r_{i,j}\!=\!\exp{(-\|\mathbf{u}_i\!-\!\mathbf{u}_j\|^{2})}$ is the repulsion weight measures similarity between denoised points, and the prior $\mathbb{P}(\mathbf{v} | \mathbf{u})$ represents the probability that a noisy point \( \mathbf{v} \) is the underlying clean point of \( \mathbf{u} \). In practice, $\mathbb{P}(\mathbf{v} | \mathbf{u})\!\propto\!\exp{(-\|\mathbf{v}\!-\!\mathbf{u}\|^{2}/(2\sigma^{2}))}$ with the distance variance $\sigma$.

\section{Experimental Results}\label{sec:Experiment}

\begin{table*}[!t]
\renewcommand{\baselinestretch}{1.0}
\renewcommand{\arraystretch}{1.0}
\setlength{\abovecaptionskip}{0pt}
\setlength{\belowcaptionskip}{0pt}
\centering
\caption{Comparison among competitive superveised denoising algorithms, CD is multiplied by $10^{5}$.}\label{sperveise:shapenet}
\begin{tabular}{@{}l|cccccccc|cccccccc@{}}
\toprule
& \multicolumn{8}{c|}{10K (Sprase)} & \multicolumn{8}{c}{30K (Dense)}\\ 
\hline
noise  & \multicolumn{2}{c}{0.01} & \multicolumn{2}{c}{0.015} & \multicolumn{2}{c}{0.02} & \multicolumn{2}{c|}{0.03} & \multicolumn{2}{c}{0.01} & \multicolumn{2}{c}{0.015} & \multicolumn{2}{c}{0.02} & \multicolumn{2}{c}{0.03} \\ 
\hline
Model & CD & EMD & CD & EMD & CD & EMD & CD & EMD & CD & EMD & CD & EMD & CD & EMD & CD & EMD \\ 
\midrule
GF & 10.99 & 2.27 & 11.01 & 2.35 & 11.66 & 2.57 & 16.78 & 3.77 & 2.95 & 2.20 & 4.84 & 3.35 & 8.21 & 5.53 & 18.03 & 12.32 \\
GLR & 5.42 & 1.05 & 6.91 & 1.43 & 9.88 & 2.12 & \textbf{11.90} & \textbf{2.70}  & 2.58 & 1.76 & 4.07 & 2.90 & 8.74 & 5.95 & 16.22 & 11.49 \\ 
\hline
PCN & 8.07 & 1.80 & 12.90 & 2.77 & 19.56 & 4.21 & 41.46 & 8.97 & 3.65 & 2.36 & 6.76 & 4.46 & 12.29 & 8.23 & 33.43 & 22.23 \\
DGCNN & 7.70 & 1.95 & 8.58 & 2.45 & 17.86 & 4.27 & 19.32 & 4.88  & 3.19 & 2.52 & 3.95 & 3.22 & 4.85 & 4.02 & 8.09 & 6.72 \\
GPD & 6.57 & 1.31 & 11.07 & 2.54 & 21.06 & 4.32 & 41.06 & 8.72 & 2.17 & 1.57 & 9.17 & 6.14 & 17.60 & 11.35 & 36.42 & 23.93 \\
TD & 10.94 & 2.61 & 15.81 & 3.59 & 22.28 & 4.96 & 40.17 & 8.71 & 3.20 & 2.41 & 5.89 & 4.04 & 11.59 & 7.67 & 28.52 & 18.95 \\
DMR & 11.19 & 2.66 & 12.92 & 2.97 & 13.70 & 3.34 & 17.43 & 4.65 & 4.39 & 3.65 & 4.91 & 4.10 & 5.77 & 4.51 & 8.25 & 6.80 \\
ScoreNet & 6.71 & 2.13 & 8.02 & 2.46 & 9.53 & 2.85 & 16.63 & 4.39 & 2.64 & 2.15 & 4.25 & 3.59 & 5.19 & 4.25 & 11.30 & 7.92 \\ 
Deeprs & 6.02 & 1.21 & 8.84 & 1.76 & 10.87 & 2.21 & 15.78 & 3.31 & 3.70 & 2.37 & 5.23 & 3.48 & 7.26 & 4.74 & 14.19 & 8.76 \\ 
PathNet & 9.99 & 2.16 & 12.82 & 2.75 & 12.98 & 2.88 & 19.11 & 4.26 & 8.14 & 5.49 & 8.45 & 6.34 & 9.66 & 7.17 & 14.70 & 10.75 \\ 
\midrule
Proposed & \textbf{4.39} & \textbf{0.90} & \textbf{6.44} & \textbf{1.37} & \textbf{8.45} & \textbf{1.84} & 13.44 & 3.40 & \textbf{2.04} & \textbf{1.46} & \textbf{2.82} & \textbf{2.08} & \textbf{3.93} & \textbf{2.93} & \textbf{7.72} & \textbf{6.67} \\ \bottomrule
\end{tabular}
\caption{Comparison among competitive unsuperveised denoising algorithms, CD  is multiplied by $10^{5}$.}\label{unsperveise:shapenet}
\begin{tabular}{@{}l|cccccccc|cccccccc@{}}
\toprule
& \multicolumn{8}{c|}{10K (Sprase)} & \multicolumn{8}{c}{30K (Dense)} \\ 
\hline
Noise & \multicolumn{2}{c}{0.01} & \multicolumn{2}{c}{0.015} & \multicolumn{2}{c}{0.02} & \multicolumn{2}{c|}{0.03} & \multicolumn{2}{c}{0.01} & \multicolumn{2}{c}{0.015} & \multicolumn{2}{c}{0.02} & \multicolumn{2}{c}{0.03} \\ 
\hline
Model & CD & EMD & CD & EMD & CD & EMD & CD & EMD & CD & EMD & CD & EMD & CD & EMD & CD & EMD \\ 
\midrule
TD & 12.15 & 2.88 & 16.57 & 3.85 & 22.30 & 5.07 & 38.05 & 8.51 & 8.99 & 6.88 & 13.61 & 10.00 & 20.02 & 14.16 & 37.84 & 26.10          \\
DMR & 11.06 & 2.79 & 12.21 & 3.40 & 14.24 & 3.53 & \textbf{17.09} & \textbf{5.10} & 6.33 & 5.14 & 7.67 & 6.45 & 9.73 & 7.05 & 16.63 & 10.12 \\
ScoreNet & 7.32 & 2.24 & 9.47 & 2.74 & 12.02 & 3.37 & 20.1 & 5.22 & 3.56 & 3.50 & 5.38 & 4.71 & 8.45 & 6.79 & 19.66 & 14.38 \\ 
\midrule
Proposed & \textbf{6.45} & \textbf{2.02}  & \textbf{8.72} & \textbf{2.50} & \textbf{11.86} & \textbf{3.21} & 21.65 & 5.36 & \textbf{3.11} & \textbf{2.72} & \textbf{4.25} & \textbf{3.46} & \textbf{6.06} & \textbf{4.60} & \textbf{12.16} & \textbf{8.53} \\ 
\bottomrule
\end{tabular}
\end{table*}

\subsection{Experimental Setting}
\subsubsection{Datasets}The training and validation set consists of post-processed point clouds sourced from the ShapeNet~\cite{chang2015shapenet} and VisionAir~\cite{Visionair} repositories. This database contains over 100 object categories described as a collection of meshes. Each selected model was sampled from over 5K points distributed uniformly and we scaled the obtained point clouds by normalizing their diameter to ensure that the data had the same statistical properties.  The training set comprises more than 20K patches of 1024 points from all categories except for those reserved for the test set. The validation set is used to monitor changes in performance metrics after the first training stage. It includes 10 point clouds belonging to five different categories, \emph{i.e.}, \emph{bath, clock, laptop, tower and train}. 

Evaluations are made on both synthetic data and real-world data to validate the proposed denoising method. 
The first type of synthetic data we used was selected from the Shapenet repository~\cite{chang2015shapenet}, where we specifically chose 100 point cloud models, each originally consisting of 30K points (dense version) from ten different categories such as \emph{airplane, bench, car, chair, lamp, pillow, rifle, sofa, speaker, and table}~\cite{pistilli2020learning}. We then downsampled these models to create a sparse version, reducing each point cloud to 10K points, to test the robustness of the denoising network. The second type of synthetic data was selected from The Stanford 3-D Scanning Repository \cite{gardner2003linear}, which includes 164 patches, each containing a varying number of points ranging from 10K to 170K. These patches cover a broad range of point densities, ensuring a uniform distribution of points across the entire dataset.
We further evaluate on the real-world \emph{Paris-rue-Madame} dataset \cite{serna2014paris} acquired using a 32-line Velodyne HDL32 LiDAR mounted on the cars. 

\subsubsection{Noise Setup} GD-GCN is evaluated under diverse noise of different levels. During training, additive Gaussian noise with the standard deviation $\sigma\in[0.01, 0.03]$ is added to the clean data to simulate different levels of noise. For test, the synthetic data are perturbed by four levels of Gaussian noise with standard deviations of 0.01, 0.015, 0.02, and 0.03, and the real-world data acquired by the high-precision vehicle-mounted laser radar is perturbed using the simulated LiDAR noise generated by the Blensor software~\cite{gschwandtner2011blensor} with the level of 0.1\% of the diagonal of bounding box.

\begin{table}[!t]
\renewcommand{\baselinestretch}{1.0}
\renewcommand{\arraystretch}{1.0}
\setlength{\tabcolsep}{1.5pt}
\setlength{\abovecaptionskip}{0pt}
\centering
\caption{Quantitative comparison among various denoising algorithms. CD  and HD  are multiplied by $10^5$ and $10^2$.}\label{supervised_unsupervised}
\begin{tabular}{@{}l|cccccc@{}}
\toprule
Mode & \multicolumn{6}{c}{Supervised/Unsupervised}  \\ 
\hline
Noise & \multicolumn{2}{c}{0.01} & \multicolumn{2}{c}{0.02} & \multicolumn{2}{c}{0.03} \\ 
\hline
Model & CD  & HD & CD & HD & CD & HD \\
\midrule
GF & 7.21 & 0.58 & 17.01 & 0.80 & 34.90 & 1.23 \\
GLR & 5.52 & 0.61 & 9.46 & 0.68 & 16.37 & 1.05 \\ 
\hline
TD & 10.12/15.04 & 0.29/0.63 & 25.32/28.49 & 0.77/0.78 & 49.61/49.15 & 1.83/1.54 \\
DMR & 5.11/8.49 & 0.20/0.54 & \textbf{7.27}/11.43  & 0.61/\textbf{0.52} & 14.62/21.63 & 0.81/0.95 \\
ScoreNet & 14.89/27.58 & 0.44/2.96 & 20.01/30.66 & 0.57/3.74 & 32.08/35.15 & 1.26/5.05 \\ 
Deeprs & 4.76 /- & 0.18 /- & 8.56/- & 0.55/- & 16.56/- & 0.89/- \\ 
PathNet & 4.80/- & 0.18/- & 16.8/- & 0.66/- & 37.6/- & 1.39/- \\ 
\hline
Proposed & \textbf{4.39}/\textbf{5.53} & \textbf{0.16}/\textbf{0.53} & 8.45/\textbf{10.55} & \textbf{0.51}/0.65 & \textbf{13.44}/\textbf{16.26} & \textbf{0.76}/\textbf{0.88} \\ 
\bottomrule
\end{tabular}
\renewcommand{\baselinestretch}{1.0}
\renewcommand{\arraystretch}{1.0}
\setlength{\abovecaptionskip}{0pt}
\setlength{\belowcaptionskip}{0pt}
\centering
\caption{Quantitative comparison among various denoising algorithms. The LiDAR noise level is set to 0.1\% of the bounding box diagonal.}\label{real_table}
\renewcommand{\baselinestretch}{1.0}
\begin{tabular}{@{}l|cccc@{}}
\toprule
Model & TD & DMR & ScoreNet & Proposed \\ 
\midrule
CD ($\times 10^{-6}$)   & 0.9  & 2.51 & 1.67     & \textbf{0.67} \\
HD ($\times 10^{-2}$)   & 0.45 & 1.13 & 0.48     & \textbf{0.19} \\
EMD   & 0.36 & 1.32 & 0.64     & \textbf{0.23} \\ 
RMSD ($\times 10^{-2}$) & 0.29 & 0.31 & 0.27   & \textbf{0.24}\\ 
\bottomrule
\end{tabular}
\end{table}

\begin{figure*}[!t]
\renewcommand{\baselinestretch}{1.0}
\setlength{\abovecaptionskip}{0pt}
\setlength{\belowcaptionskip}{0pt}
\centering
\includegraphics[width=0.95\textwidth]{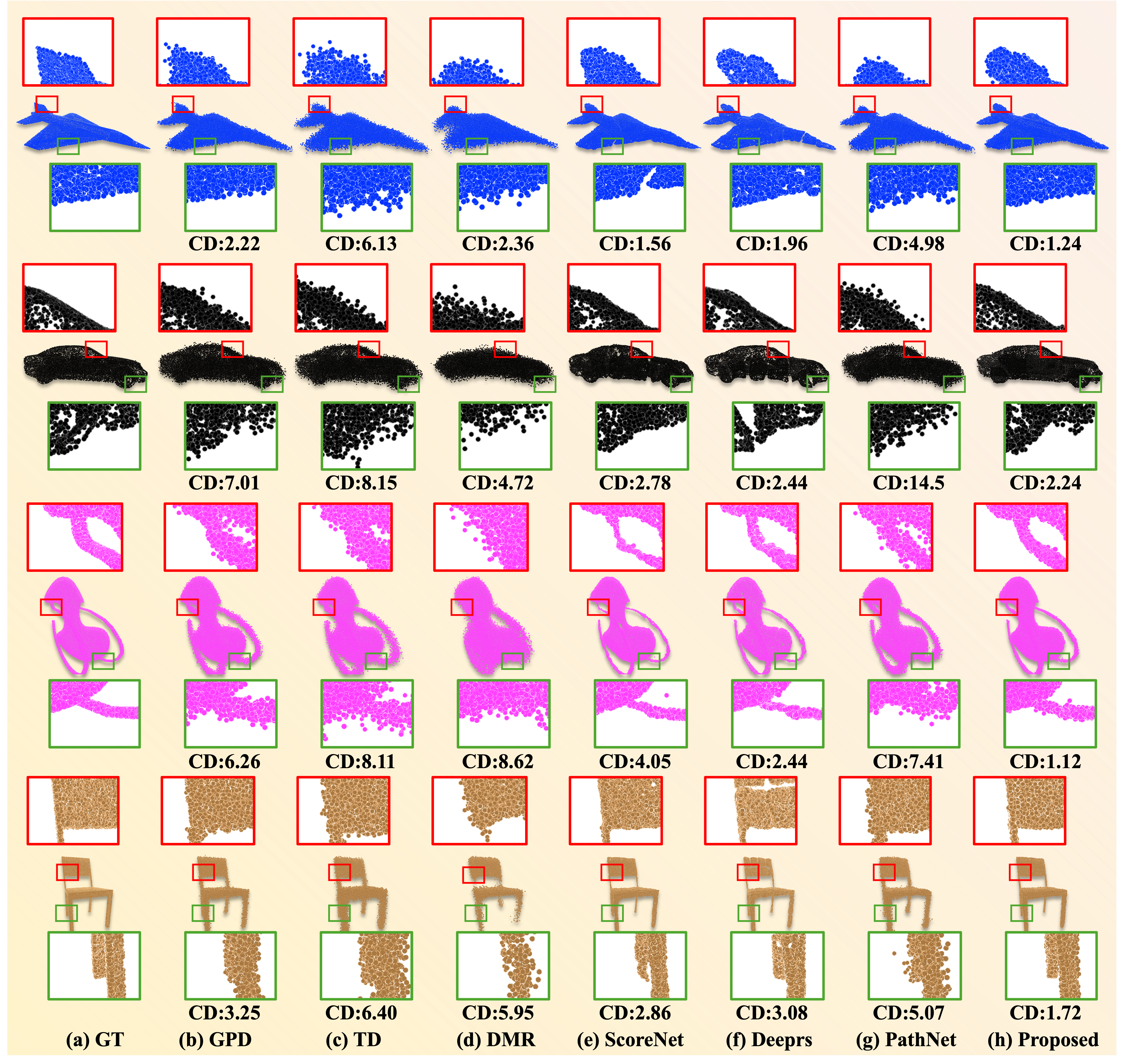}
\caption{
Denoising results for \emph{airplane} and \emph{car} from the ShapeNet repository. 
From top left to bottom right: Groud Truth, GPDNet~\cite{pistilli2020learning}, TD~\cite{hermosilla2019total}, DMR~\cite{luo2020differentiable}, ScoreNet~\cite{luo2021score}, Deeprs~\cite{chen2022deep}, and PathNet~\cite{wei2024pathnet} and the proposed method. CD values in $10^{-6}$ are reported.
}\label{shapenet_data}
\end{figure*}

\subsubsection{Baselines} We adopt ten baselines for comparison, including two optimization-based methods (\emph{i.e.}, Guided filtering (GF)~\cite{he2010guided} and GLR~\cite{zeng20193d}) and eight learning-based methods (\emph{i.e.}, PCN (PointCleanNet)~\cite{rakotosaona2020pointcleannet}, TD (TotalDenoising)~\cite{hermosilla2019total}, DGCNN~\cite{wang2019dynamic}, GPDNet~\cite{pistilli2020learning}, DMR~\cite{luo2020differentiable}, ScoreNet~\cite{luo2021score}, Deeprs~\cite{chen2022deep}, and PathNet~\cite{wei2024pathnet}). GF~\cite{he2010guided} is an extended version of the famous guided image filtering method for point cloud denoising, whereas GLR~\cite{zeng20193d} uses the patch manifold prior with graph regularization term. DGCNN~\cite{wang2019dynamic} is originally developed for point cloud segmentation and classification and is elegantly modified to adapt to the denoising task. PCN~\cite{rakotosaona2020pointcleannet} and TD~\cite{hermosilla2019total} are classical learning-based methods, while GPDNet~\cite{pistilli2020learning}, DMR~\cite{luo2020differentiable}, ScoreNet~\cite{luo2021score}, Deeprs~\cite{chen2022deep}, and PathNet~\cite{wei2024pathnet} are most recent high-performance denoising networks whose official codes are publicly available. We train PCN~\cite{rakotosaona2020pointcleannet}, GPDNet~\cite{pistilli2020learning}, and TD~\cite{hermosilla2019total} using the same optimal training parameters as reported in their papers, and train the remaining models to achieve best performance.

\begin{figure*}[!t]
\renewcommand{\baselinestretch}{1.0}
\centering
\setlength{\abovecaptionskip}{0pt}
\setlength{\belowcaptionskip}{0pt}
\includegraphics[width=0.9\linewidth]{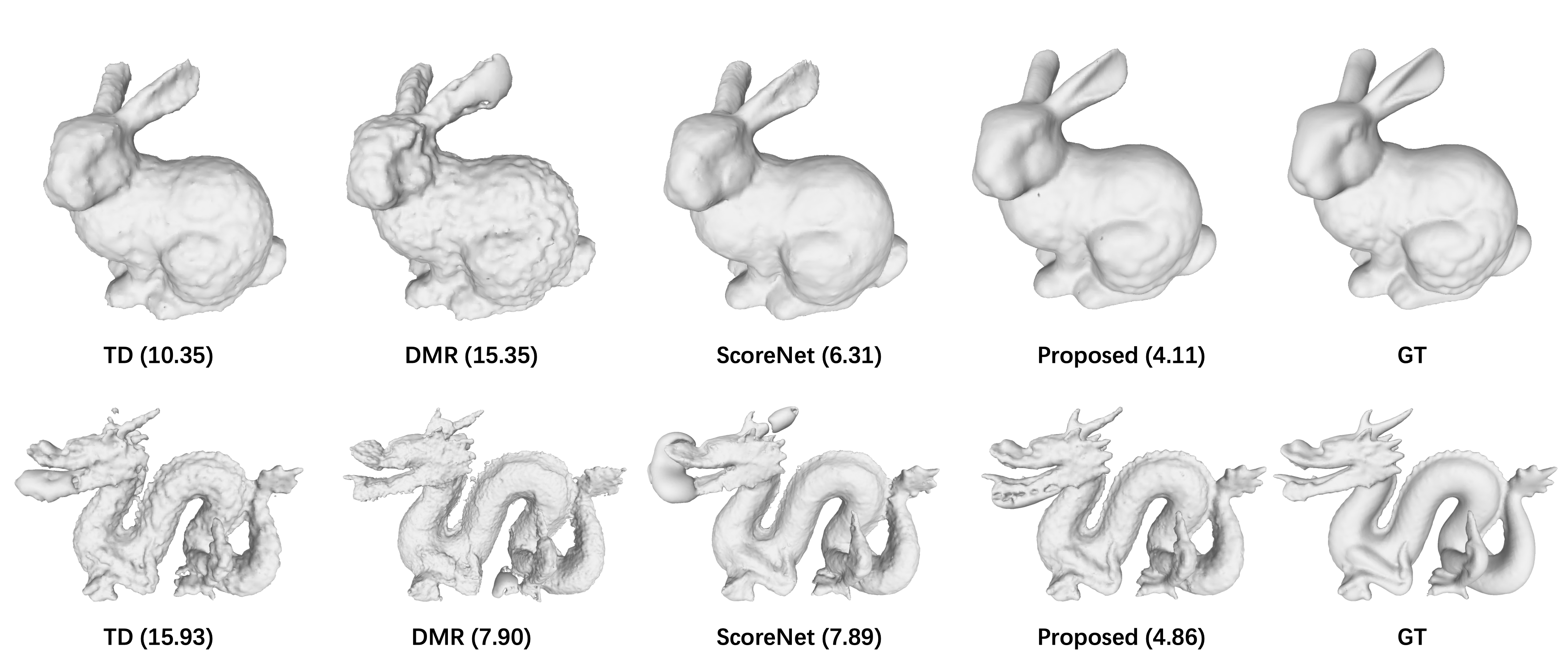}
\caption{Denoising results for the \emph{bunny} and \emph{dragon} from the Stanford repository using both supervised (top) and unsupervised (bottom) modes. From top left to bottom right: TD~\cite{hermosilla2019total}, DMR~\cite{luo2020differentiable}, ScoreNet~\cite{luo2021score}, ours and ground truth. CD values in $10^{-6}$ are reported.
}\label{standford_data}
\includegraphics[width=0.9\linewidth]{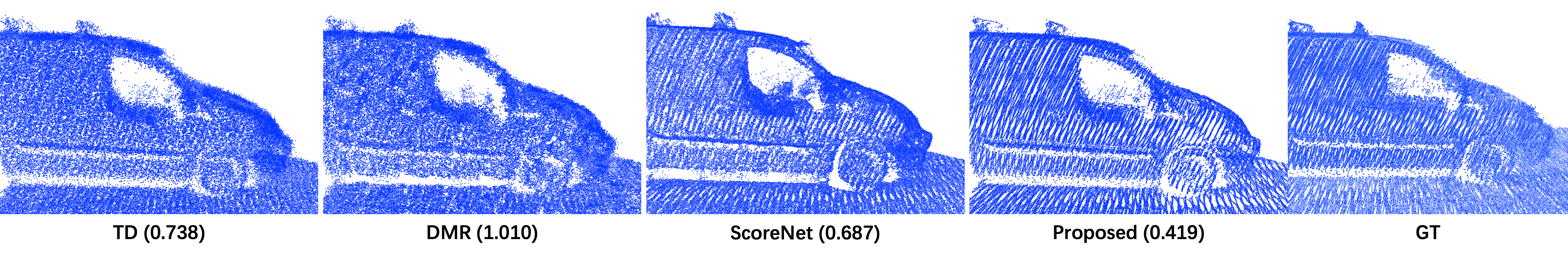}
\caption{Denoising real data results for real-world data with LiDAR noise. From left to right: TD~\cite{hermosilla2019total}, DMR~\cite{luo2020differentiable}, ScoreNet~\cite{luo2021score}, ours and ground truth. The numbers in parentheses are CD  values, which are multiplied by $10^{6}$.}
\label{real_experiment}
\end{figure*}

\subsubsection{Metrics} We adopt the widely used metrics for denoising, including Chamfer distance~(CD), Earth Mover’s distance (EMD), and Hausdorff distance~(HD) between the clean point cloud $\mathcal{U}$ and its denoised version $\hat{\mathcal{U}}$. CD and EMD are formalized in~\eqref{CD distance} and~\eqref{EMD distance}. HD measures the similarity between $\hat{\mathcal{U}}$ and $\mathcal{U}$. 
\begin{equation}\label{hd distance}
d_{\rm H}( \hat{\mathcal{U}} , \mathcal{U})=\max \left\{\sup_{\hat{u} \in \hat{\mathcal{U}}} d(\hat{u}, \mathcal{U}), \sup_{u \in \mathcal{U}} d(\hat{\mathcal{U}}, u)\right\},
\end{equation}
where $d(\hat{u}, \mathcal{U})=\inf_{u \in \mathcal{U}}d(\hat{u},u)$ is the minimum distance from the point $\hat{u}\in\hat{\mathcal{U}}$ to $\mathcal{U}$ and $d(\hat{\mathcal{U}}, u)=\inf_{\hat{u} \in \hat{\mathcal{U}}}d(\hat{u},u)$ the minimum distance from $u\in\mathcal{U}$ to $\hat{\mathcal{U}}$. HD is sensitive to the boundary and can eliminate unreasonable distances caused by outliers. For real-word data with non-uniformly distributed points, we also adopt the root mean square value of the surface distance (RMSD) to measure the point-surface distance.
\begin{equation}\label{rmsd distance}
d_{\rm RMSD}( \hat{\mathcal{U}} , \mathcal{U})=\sqrt{\frac{1}{N} \sum_{i=1}^{N} \min _{j}\left\|\hat{\mathbf{u}}_{i}-\mathbf{u}_{j}\right\|_{2}^{2}}.
\end{equation}


\subsubsection{Hyper-parameters}
We fix the hyperparameters to train one single model for all the experiments, except for ablation studies. GD-GCN is trained under a fixed noise variance for 1M iterations using a batch size of 8. We adopt the Adam optimizer with learning rate scheduler \emph{ReduceLROnPlateau}, where the initial learning rate is $10^{-3}$ and the minimum learning rate is $10^{-6}$. $k$ is set to 16 for the $k$NN algorithm. The termination time of the integral $T$ in~\eqref{time window GCN} is set to 1.

\subsection{Quantitative Results}
Table~\ref{sperveise:shapenet} reports the CD and EMD metrics for the supervised denoising method at four different noise levels, averaged over the first type of synthetic test data.
The values filled in are averaged over all the synthetic test point cloud models. 
In Table~\ref{sperveise:shapenet}, the left half shows the results on the sparse version of the test set, and the right half shows the results on the dense version. GD-GCN outperforms the previous best performing method on both the sparse and dense versions in most cases, achieving significant reductions in both metrics. The only exception is the 3\% noise condition on the sparse version, where it is slightly worse than the GLR~\cite{zeng20193d} model. 
As shown in Table~\ref{unsperveise:shapenet}, we select three models that have been experimentally verified and performed well for the comparison of unsupervised point cloud denoising performance with GD-GCN. The unsupervised version of GD-GCN is not as good as its supervised counterpart, but it still beats TD~\cite{hermosilla2019total} and ScoreNet~\cite{luo2021score} at all noise levels. The only exception is the 3\% noise condition on the sparse version of the first type test set, where our unsupervised version is slightly worse.

\begin{figure*}[!t]
\renewcommand{\baselinestretch}{1.0}  
\setlength{\abovecaptionskip}{0pt}
\setlength{\belowcaptionskip}{0pt}
\centering
\includegraphics[width=\linewidth]{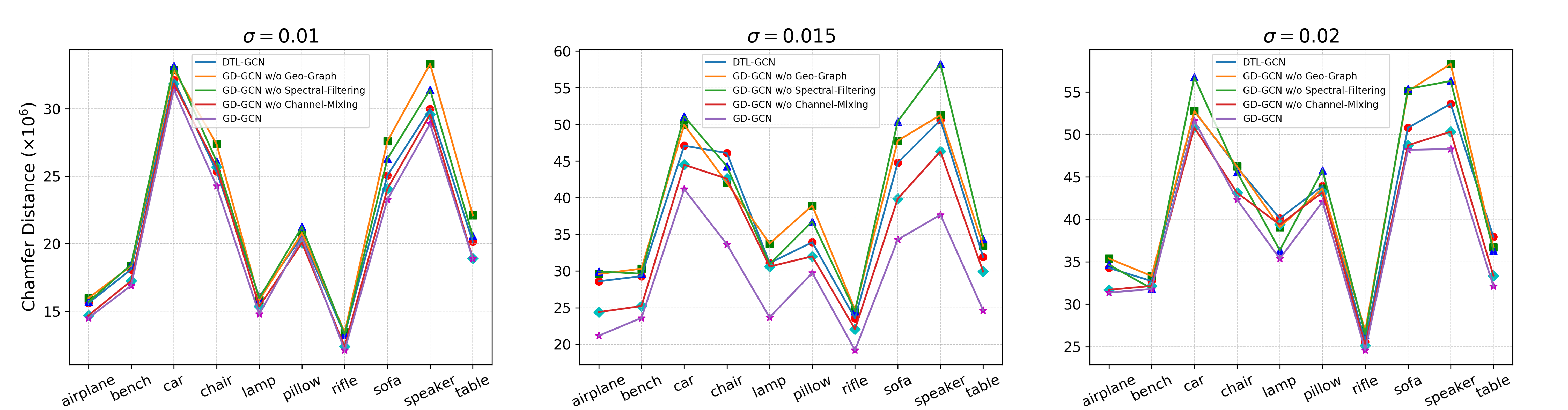}
\caption{Ablation experiment results.  The performance curves of DTL-GCN, GD-GCN w/o Geo-Graph, GD-GCN w/o Spectral-Filtering, and Proposed in the ablation experiment, conducted under noise levels of $1\%$, $1.5\%$, and $2\%$. }\label{ablation_experiment}
\end{figure*}

We also evaluate denosing performance in terms of CD and HD on the other synthetic test set. 
Table~\ref{supervised_unsupervised} compares GD-GCN with five other models: two optimization models GF~\cite{he2010guided} and GLR~\cite{zeng20193d}, and five learning models TD~\cite{hermosilla2019total}, DMR~\cite{luo2020differentiable} ScoreNet~\cite{luo2021score}, Deeprs~\cite{chen2022deep}, and PathNet~\cite{wei2024pathnet}.
The supervised version of CD does not achieve the best performance under the 2\% noise level. The same applies to the unsupervised version of HD under the same noise level. Apart from these cases, GD-GCN consistently yields state-of-the-art performance in both supervised or unsupervised modes.  

Furthermore, we verify the generalization ability of GD-GCN using real-world LiDAR data that is not subjected to any normalization pre-processing.
Three representative models, \emph{i.e.}, TD~\cite{hermosilla2019total}, DMR~\cite{luo2020differentiable}, and ScoreNet~\cite{luo2021score}, are selected for comparison and the RMSD metric is adopted in addition to CD, EMD, and HD.
Table~\ref{real_table} shows that GD-GCN outperforms all the three models under all the metrics. 




\subsection{Qualitative Results}
We visualize in \figurename~\ref{shapenet_data}--\figurename~\ref{real_experiment} the denoised point clouds from synthetic and real-world datasets. Here, we only consider GD-GCN and six recent denoising networks (\emph{i.e.}, GPD~\cite{pistilli2020learning}, DMR~\cite{luo2020differentiable}, ScoreNet~\cite{luo2021score}, Deeprs~\cite{chen2022deep}, and PathNet~\cite{wei2024pathnet}) that obtain satisfactory quantitative results for visualization. 

\figurename~\ref{shapenet_data} presents denoised \emph{airplane}, \emph{car}, and two \emph{chairs} from the ShapeNet repository under 1\% Gaussian noise. 
Compared with the leftmost ground truth (clean) point cloud, GPD~\cite{pistilli2020learning}, TD~\cite{hermosilla2019total}, DMR~\cite{luo2020differentiable}, and PathNet~\cite{wei2024pathnet} exhibit significant noise and cannot accurately fit the underlying surface. ScoreNet~\cite{luo2021score} and Deeprs~\cite{chen2022deep} use gradient ascent scores in an iterative approach to approximate the underlying surface and remove noise but cause structural loss in the shape.

\figurename~\ref{standford_data} shows the denoising results for \emph{bunny} and \emph{dragon} from the 3-D Stanford repository under 1\% Gaussian noise. Both objects have over 100K points. We meshed their denoised results to facilitate a comparison of their performance.
As shown in the top row under supervised mode, all models can recover the contour of the \emph{bunny}. However, our model produces smoother and more refined denoising results that closely approximate the underlying surface of the object. In contrast, as shown in the bottom row under unsupervised mode, the denoising results are generally inferior to those achieved with supervised learning. Nevertheless, our method still preserves the structure of the \emph{dragon} most accurately.

\figurename~\ref{real_experiment} shows the denoising results for the real-world dataset \emph{Paris-rue-Madame}. Despite the non-uniform distribution of the point clouds acquired by laser radar, our denoised result is cleaner and smoother than that of  TD~\cite{hermosilla2019total} and DMR, and retains more details than ScoreNet~\cite{luo2021score}.

\subsection{Ablation Studies}
Ablation studies are performed to validate the main designs of the proposed method.
\subsubsection{Continuous Convolution vs. Discrete Convolution} 
We compare the proposed method with a baseline named DTL-GCN that uses three discrete integer-step graph convolution layers with each followed by batch normalization and nonlinear activation, as formulated below.
\begin{equation}\label{DTL-GCN}
\mathbf{Z}^{l+1} = \mathbf{A}_{\mathcal{G}}^{l}\mathbf{Z}^{l+1}\Theta^{l}.
\end{equation}

\subsubsection{Geometric Graph vs. Euclidean Graph} 
We also compare our model with the second baseline that adopt dynamic graph construction with Euclidean metric.
The adjacency matrix $\tilde{\mathbf{A}}_{\mathcal{G}}$ using the Euclidean metric is
\begin{equation}
\tilde{a}_{ij} =\exp{\left(-\frac{\left \| \mathbf{u}_{i}- \mathbf{u}_{j} \right \|^{2}}{2\tilde{\delta}^{2}}\right)}, j\in \mathcal{N}(i),
\label{fixed adjacency matrix}
\end{equation}
where \(\tilde{\delta}\) denotes the standard deviation of the Euclidean distance of each point to its neighbors. Apart from this modification, the rest of the network remains consistent with proposed GD-GCN. The second model is referred to as GD-GCN w/o Geo-Graph.

\subsubsection{Proposed Graph Spectral Filtering vs. Graph Neural Diffusion} 
We further compare our model with the third baseline that replace the  graph spectral filtering in~\eqref{bernstein polynomial} with graph diffusion kernel~\cite{chamberlain2021beltrami}. We use an improved version of the graph diffusion kernel, ${\mathbf{A}}^{t}_{\mathcal{G}}$, because the edge weights between nodes in ${\mathbf{A}}^{t}_{\mathcal{G}}$ are set by considering the geometric relationships among the nodes as described in Section~\ref{sec:metric learning}. The third baseline is denoted as GD-GCN w/o Spectral-Filtering. Furthermore, we also validated a configuration where only the channel mixing matrices were removed, while the other settings remained consistent with the proposed GD-GCN. This fourth baseline is referred to as GD-GCN w/o Channel-Mixing.

\begin{figure}[!t]
\renewcommand{\baselinestretch}{1.0}  
\setlength{\abovecaptionskip}{0pt}
\setlength{\belowcaptionskip}{0pt}
\centering
\includegraphics[width=1\linewidth]{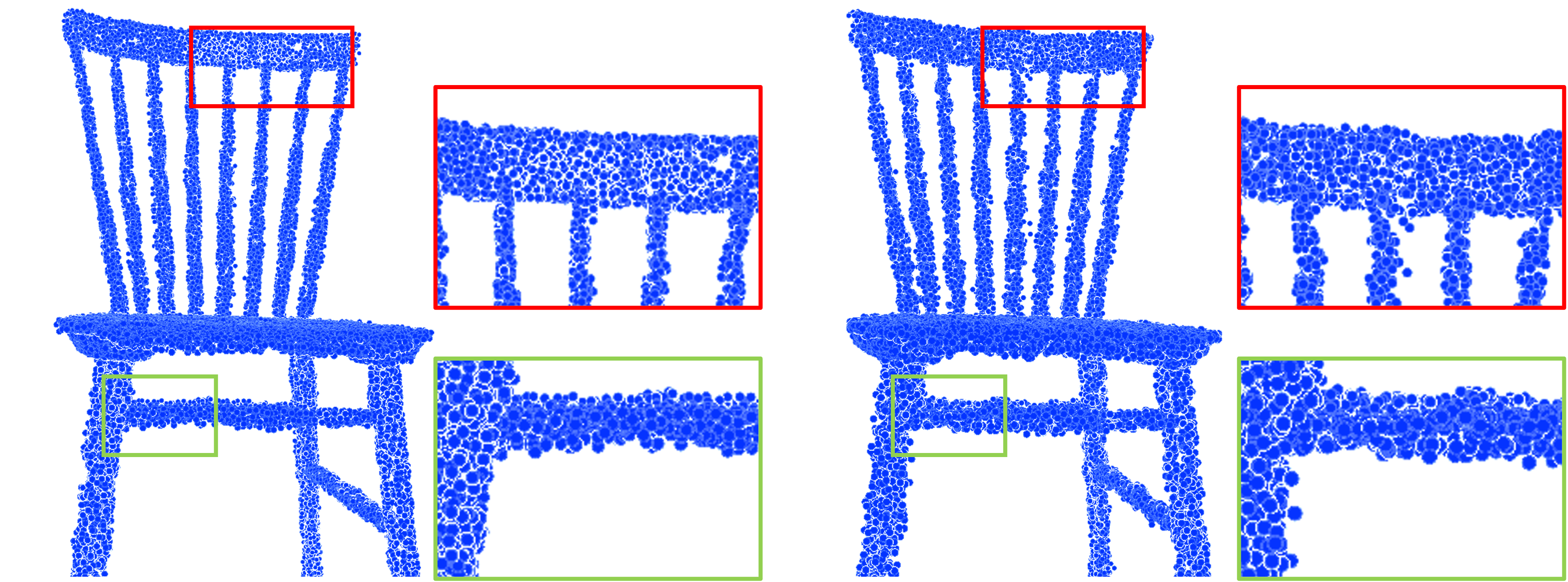}
\caption{Outcomes of the denoising process achieved through geometric graph construction with Riemannian metric~(left, CD:$1.550\times 10^{-6}$) and graph construction based on Euclidean metric~(right, CD: $1.637\times 10^{-6}$). 
}\label{geometric graph construction compare}
\end{figure}

We evaluate the proposed GD-GCN and four baselines based on point clouds of ten categories with 1.0\%, 1.5\%, and 2.0\% additive Gaussian noise in the 30K test set. We use the Chamfer distance (CD) to measure the denoising performance of several models. As shown in \figurename~\ref{ablation_experiment}, GD-GCN exhibits exceptional denoising performance across different levels of noise and for various types of objects. Fig.~\ref{geometric graph construction compare} demonstrates the difference between the geometric graph (GD-GCN) and Euclidean graph model (GD-GCN w/o Geo-Graph) in details in the denoised point clouds. GD-GCN obtains clearer and more realistic object edges, whereas the edges generated by the GD-GCN w/o Geo-Graph model appear blurred.

\begin{figure*}[!t]
\renewcommand{\baselinestretch}{1.0}
\setlength{\abovecaptionskip}{0pt}
\setlength{\belowcaptionskip}{0pt}
\centering
\includegraphics[width=0.9\linewidth]{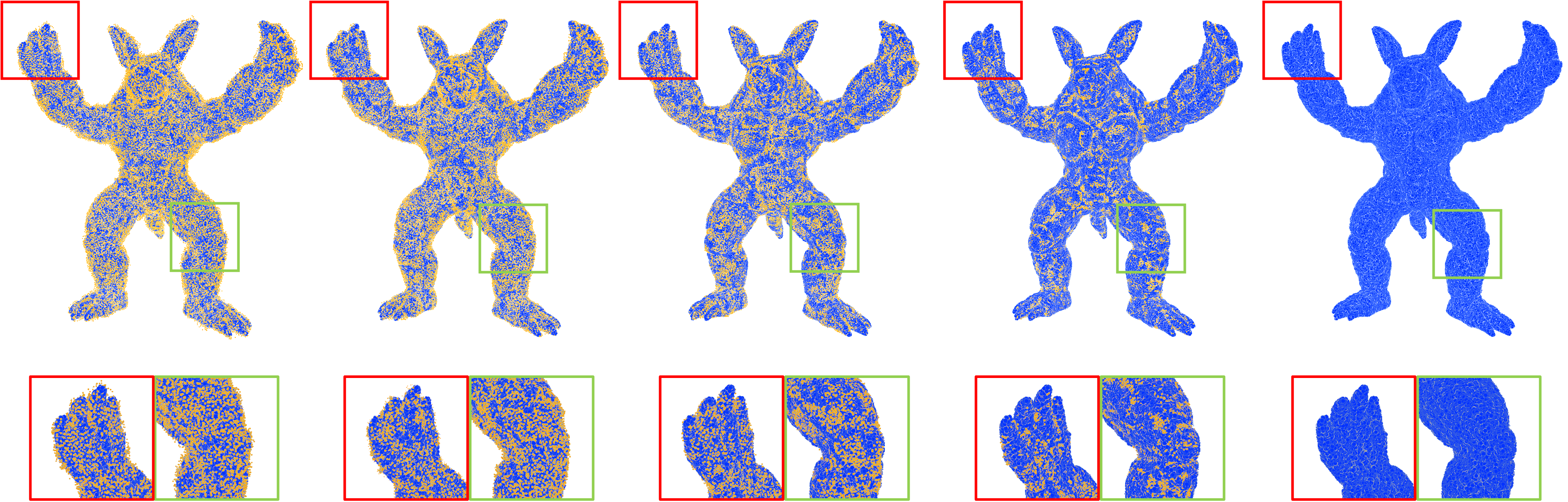}
\caption {Dynamic denoising analysis of MST-GConv. Different termination times lead to Different denoising results, where orange represents denoised points and blue represents true underlying points. From left to right: 0.25T~(9.437), 0.5T~(6.444), 0.75T~(3.824), T~(2.679), and the ground truth. The numbers in parentheses are CD  values, which are multiplied by $10^{-6}$.}\label{gradual_change}
\end{figure*}

\subsubsection{Dynamic denoising analysis of MST-GConv}
As described in Section~\ref{sec:MST-GConv}, MST-GConv is essentially a dynamic system with a numerical solution that can be obtained through neural PDEs, as shown in~\eqref{FPTV GCN}. Both the input and output exist in the same dimensional feature space in MST-GConv. Therefore, we can extract the hidden states of MST-GConv at any time step from $0 \sim T$ and connect them to the rest coordinate reconstruction. By comparing the resulting denoising outputs, we can demonstrate the feature learning process of MST-GConv in the latent space.

\figurename~\ref{gradual_change} depicts a well-trained neural network in which we extract the intermediate states of MST-GConv at specific time points of $0.25T$, $0.5T$, $0.75T$, and $T$. We then feed these extracted states into a coordinate reconstruction model that has also undergone training, resulting in denoised outputs. As time increases in~\eqref{FPTV GCN}, the hidden state of MST-GConv changes, leading to a gradual approach of the final denoised point cloud to the surface of the real object.

\section{Conclusions}\label{sec:con}
In this paper, we propose a novel fine-granularity dynamic graph convolution network (GD-GCN) that directly consumes noisy 3-D point clouds. We generalise existing graph convolution networks to dynamic systems through a micro-step temporal graph convolution (MST-GConv) module, which enables fine-grained feature learning and topology evolution. The geometric graph construction is implemented by learning a \emph{Riemannian metric}, so that the similarity between nearby points is calculated based on Riemannian distance. This can better capture the local shape properties. We also propose a robust graph spectral filtering method that is formed by Bernstein polynomial approximation. It bounds the filtered eigenvalues and guarantees the well-posedness of MST-GConv, while making fractional precision graph convolution more flexible with arbitrary spectrum response. Extensive experiments demonstrate the superiority of GD-GCN compared to its simplified counterparts and a significant improvement over state-of-the-art methods.

\appendices
\section{Proof of Proposition~\ref{pro:1}}\label{app:1}



We first prove the existence of coefficients that satisfy the condition that the value of $\mathbf{B_{K}}(\lambda)$ is positive. The $k$-th Bernstein basis $b_{k}^{K}(\lambda)$ is obviously positive for $\binom{K}{k}=\frac{K!}{k!(K-k)!}>0$, $(1-\lambda)^{K-k}>0$, and $\lambda^{k}>0$.
Therefore, $\mathbf{B_{K}}(\lambda)$ is non-negative as long as the coefficient $\theta_k$ is positive.

We then prove that the upper bound of $\mathbf{B_{K}}(\lambda)$ can be mapped to 1 by normalizing the coefficients.
The derivative of $b_{k}^{K}(\lambda)$ is:
\begin{align}\label{eq:derivative1}
\frac{\mathrm{d}}{\mathrm{d}\lambda} b^{K}_{k}\!(\lambda)\! &=\!\frac{\mathrm{d}}{\mathrm{d}\lambda}\!\left[\binom{K}{k} \lambda^{k}(1\!-\!\lambda)^{K\!-\!k}\right]\!=\!\binom{K}{k}\!\left[\frac{\mathrm{d}}{\mathrm{d}\lambda} \lambda^{k}(1\!-\!\lambda)^{K\!-\!k}\right] \nonumber\\ 
& =\! \binom{K}{k}\left[k \lambda^{k\!-\!1}(1\!-\!\lambda)^{K\!-\!k}\!-\!(K\!-\!k)\lambda^{k}(1\!-\!\lambda)^{K\!-\!k\!-\!1}\right] \nonumber\\ 
& = \binom{K}{k}\lambda^{k-1}(1-\lambda)^{K-k-1}[k(1-\lambda)-(K-k) \lambda] \nonumber\\ & = \binom{K}{k}\lambda^{k-1}(1-\lambda)^{K-k-1}(k-K\lambda).
\end{align}
$b^{K}_{k}(\lambda)$ reaches its maximum value at $\lambda =k/K$ where the derivatives $\mathrm{d}b^{K}_{k}(\lambda)/\mathrm{d}\lambda=0$. We find $k$ by considering the derivative of $\log b_k^K(\lambda)$ for $\lambda=k/K$ using~\eqref{eq:derivative1}. 
\begin{align}
&\frac{\mathrm{d}}{\mathrm{d}k} \log \left(b^{K}_{k}\left(\frac{k}{K}\right)\right) \nonumber\\
&\quad=\frac{\mathrm{d}}{\mathrm{d}k}\left[\log \binom{K}{k} + k \log \left(\frac{k}{K}\right)+(K-k) \log \left(1-\frac{k}{K}\right)\right]  \nonumber\\
&\quad=\frac{\mathrm{d}}{\mathrm{d}k}
\binom{K}{k} + \log \left(\frac{k}{K}\right)-\log \left(1-\frac{k}{K}\right)\nonumber\\
&\quad=\frac{\mathrm{d}}{\mathrm{d}k}\binom{K}{k} + \log \left(\frac{k}{K-k}\right) 
\end{align}

Note that $\binom{K}{k}$ is logarithmic convex but not continuous with regard to $k$. Thus, $b_{k}^{K}(\frac{k}{K})$ achieves the maximum value when $k=\lfloor K/2\rfloor$, despite that $\frac{\mathrm{d}}{\mathrm{d}k}\binom{K}{k}$ does not exist.
\begin{equation}
\max_{x \in[0,1)} b_{k}^{K}\left(\frac{k}{K}\right) \leq b_{\lfloor \frac{K}{2}\rfloor}^{K}\left(\frac{1}{2}\right)= 2^{-K}\binom{K}{\lfloor \frac{K}{2}\rfloor}.
\end{equation}
The upper bound is formulated as:
\begin{equation}
\max_{x\in[0,1)}\!\mathbf{B_{K}}(\lambda)\!<\!\sum_{k=0}^{K}\!\theta_{k}\!\max_{\lambda \in[0,1)}\!b_{k}^{K}(\lambda)\!=\!2^{-K}\!\sum_{k=0}^{K}\!\theta_{k}\!\binom{K}{\lfloor\!\frac{K}{2}\!\rfloor}. 
\end{equation}
Therefore, the maximum value of $\mathbf{B_{K}}(\lambda)$ becomes 1 by normalizing the learned parameters with the normalization factor $\frac{(K-\lfloor\frac{K}{2}\rfloor)!}{(K+1)!}\cdot2^{K}$.


\begin{thebibliography}{10}

\providecommand{\newblock}{\relax}
\providecommand{\bibinfo}[2]{#2}
\providecommand{\BIBentrySTDinterwordspacing}{\spaceskip=0pt\relax}
\providecommand{\BIBentryALTinterwordstretchfactor}{4}
\providecommand{\BIBentryALTinterwordspacing}{\spaceskip=\fontdimen2\font plus
\BIBentryALTinterwordstretchfactor\fontdimen3\font minus
  \fontdimen4\font\relax}
\providecommand{\BIBforeignlanguage}[2]{{%
\expandafter\ifx\csname l@#1\endcsname\relax
\typeout{** WARNING: IEEEtran.bst: No hyphenation pattern has been}%
\typeout{** loaded for the language `#1'. Using the pattern for}%
\typeout{** the default language instead.}%
\else
\language=\csname l@#1\endcsname
\fi
#2}}
\providecommand{\BIBdecl}{\relax}
\BIBdecl





\bibitem{9438625}
Z.~Yuan, X.~Song, L.~Bai, Z.~Wang, and W.~Ouyang, ``Temporal-channel transformer for 3D Lidar-based video object detection for autonomous driving,'' \emph{IEEE Trans. Circuits Syst. Video Technol.}, vol.~32, no.~4, pp.~2068--2078, 2022.

\bibitem{9504551}
L.~Zhao, J.~Guo, D.~Xu, and L.~Sheng, ``Transformer3D-Det: Improving 3D object detection by vote refinement,'' \emph{IEEE Trans. Circuits Syst. Video Technol.}, vol.~31, no.~12, pp. 4735--4746, 2021.

\bibitem{9810920}
H.~Sheng, R.~Cong, D.~Yang, R.~Chen, S.~Wang, and Z.~Cui, ``UrbanLF: A comprehensive light field dataset for semantic segmentation of urban scenes,'' \emph{IEEE Trans. Circuits Syst. Video Technol.}, vol.~32, no.~11, pp. 7880--7893, 2022.

\bibitem{9496619}
P.~Zhang, X.~Wang, L.~Ma, S.~Wang, S.~Kwong, and J.~Jiang, ``Progressive point cloud upsampling via differentiable rendering,'' \emph{IEEE Trans. Circuits Syst. Video Technol.}, vol.~31, no.~12, pp. 4673--4685, 2021.

\bibitem{9493165}
D.~Ding, C.~Qiu, F.~Liu, and Z.~Pan, ``Point cloud upsampling via perturbation learning,'' \emph{IEEE Trans. Circuits Syst. Video Technol.}, vol.~31, no.~12, pp. 4661--4672, 2021.

\bibitem{digne2017bilateral}
J.~Digne and C.~De~Franchis, ``The bilateral filter for point clouds,'' \emph{Image Process. On Line}, vol.~7, pp. 278--287, 2017.

\bibitem{zhang2019point}
F.~Zhang, C.~Zhang, H.~Yang, and L.~Zhao, ``Point cloud denoising with principal component analysis and a novel bilateral filter,'' \emph{Traitement du Signal}, vol.~36, no.~5, pp. 393--398, 2019.

\bibitem{alexa2003computing}
M.~Alexa, J.~Behr, D.~Cohen-Or, S.~Fleishman, D.~Levin, and C.~T. Silva, ``Computing and rendering point set surfaces,'' \emph{IEEE Trans. Vis. Comput. Graphics}, vol.~9, no.~1, pp. 3--15, 2003.

\bibitem{oztireli2009feature}
A.~C. {\"O}ztireli, G.~Guennebaud, and M.~Gross, ``Feature preserving point set surfaces based on non-linear kernel regression,'' \emph{Comput. Graph. Forum}, vol.~28, no.~2, pp. 493--501, 2009.

\bibitem{guennebaud2007algebraic}
G.~Guennebaud and M.~Gross, ``Algebraic point set surfaces,'' \emph{ACM Trans. Graph.}, vol.~26, no.~3, article no. 23, pp. 1--10, 2007.

\bibitem{lipman2007parameterization}
Y.~Lipman, D.~Cohen-Or, D.~Levin, and H.~Tal-Ezer, ``Parameterization-free projection for geometry reconstruction,'' \emph{ACM Trans. Graph.}, vol.~26, no.~3, article no. 22, pp. 1--6, 2007.

\bibitem{huang2013edge}
H.~Huang, S.~Wu, M.~Gong, D.~Cohen-Or, U.~Ascher, and H.~Zhang, ``Edge-aware point set resampling,'' \emph{ACM Trans. Graph.}, vol.~32, no.~1, article no. 9, pp. 1--12, 2013.

\bibitem{cazals2005estimating}
F.~Cazals and M.~Pouget, ``Estimating differential quantities using polynomial fitting of osculating jets,'' \emph{Comput. Aided Geom. Des.}, vol.~22, no.~2, pp. 121--146, 2005.

\bibitem{avron2010l1}
H.~Avron, A.~Sharf, C.~Greif, and D.~Cohen-Or, ``$\ell_1$-sparse reconstruction of sharp point set surfaces,'' \emph{ACM Trans. Graph.}, vol.~29, no.~5, article no. 135, pp. 1--12, 2010.

\bibitem{sun2015denoising}
Y.~Sun, S.~Schaefer, and W.~Wang, ``Denoising point sets via $L_0$ minimization,'' \emph{Comput. Aided Geom. Des.}, vol.~35, pp. 2--15, 2015.

\bibitem{mattei2017point}
E.~Mattei and A.~Castrodad, ``Point cloud denoising via moving RPCA,'' \emph{Comput. Graph. Forum}, vol.~36, no.~8, pp. 123--137, 2017.

\bibitem{zeng20193d}
J.~Zeng, G.~Cheung, M.~Ng, J.~Pang, and C.~Yang, ``{3D} point cloud denoising using graph laplacian regularization of a low dimensional manifold model,'' \emph{IEEE Trans. Image Process.}, vol.~29, pp. 3474--3489, 2019.

\bibitem{dinesh20183d}
C.~Dinesh, G.~Cheung, and I.~V. Bajic, ``{3D} point cloud denoising via bipartite graph approximation and reweighted graph {Laplacian},'' \emph{arXiv preprint arXiv:1812.07711}, 2018.

\bibitem{schoenenberger2015graph}
Y.~Schoenenberger, J.~Paratte, and P.~Vandergheynst, ``Graph-based denoising for time-varying point clouds,'' in \emph{2015 3DTV-Conf.: True Vis. - Capture, Transmiss. Display 3D Video (3DTV-CON)}, 2015.

\bibitem{rakotosaona2020pointcleannet}
M.-J. Rakotosaona, V.~La~Barbera, P.~Guerrero, N.~J. Mitra, and M.~Ovsjanikov, ``PointCleanNet: Learning to denoise and remove outliers from dense point clouds,'' \emph{Comput. Graph. Forum}, vol.~39, no.~1, pp. 185--203, 2020.

\bibitem{duan20193d}
C.~Duan, S.~Chen, and J.~Kovacevic, ``3D point cloud denoising via deep neural network based local surface estimation,'' in \emph{2019 IEEE Int. Conf. Acoust., Speech Signal Process. (ICASSP)}, 2019, pp. 8553--8557.

\bibitem{hermosilla2019total}
P.~Hermosilla, T.~Ritschel, and T.~Ropinski, ``Total denoising: Unsupervised learning of {3D} point cloud cleaning,'' in \emph{2019 IEEE/CVF Int. Conf. Comput. Vis. (ICCV)}, 2019, pp. 52--60.

\bibitem{luo2020differentiable}
S.~Luo and W.~Hu, ``Differentiable manifold reconstruction for point cloud denoising,'' in \emph{Proc. 28th ACM Int. Conf. Multimedia}, 2020, pp. 1330--1338.

\bibitem{luo2021score}
------, ``Score-based point cloud denoising,'' in \emph{2021 IEEE/CVF Int. Conf. Comput. Vis. (ICCV)}, 2021, pp. 4583--4592.

\bibitem{dinesh20193d}
C. Dinesh, G. Cheung, and I. Bajić, ``3D point cloud color denoising using convex graph-signal smoothness priors,'' in \emph{2019 IEEE 21st Int. Workshop Multimedia Signal Process. (MMSP)}, 2019, pp. 1--6.

\bibitem{qi2017pointnet}
C.~R. Qi, H.~Su, K.~Mo, and L.~J. Guibas, ``PointNet: Deep learning on point sets for 3D classification and segmentation,'' in \emph{2017 IEEE Conf. Comput. Vis. Pattern Recognit.}, 2017, pp. 652--660.

\bibitem{pistilli2020learning}
F.~Pistilli, G.~Fracastoro, D.~Valsesia, and E.~Magli, ``Learning graph-convolutional representations for point cloud denoising,'' in \emph{16th Eur. Conf. Comput. Vis. (ECCV)}, 2020, pp. 103--118.

\bibitem{dgcnn}
Y.~Wang, Y.~Sun, Z.~Liu, S.~E. Sarma, M.~M. Bronstein, and J.~M. Solomon, ``Dynamic graph CNN for learning on point clouds,'' \emph{ACM Trans. Graph.}, vol.~38, no.~5, article no. 146, pp. 1--12, 2019.

\bibitem{bronstein2017geometric}
M.~M. Bronstein, J.~Bruna, Y.~LeCun, A.~Szlam, and P.~Vandergheynst, ``Geometric deep learning: Going beyond Euclidean data,'' \emph{IEEE Signal Process. Mag.}, vol.~34, no.~4, pp. 18--42, 2017.

\bibitem{shuman2013emerging}
D.~I. Shuman, S.~K. Narang, P.~Frossard, A.~Ortega, and P.~Vandergheynst, ``The emerging field of signal processing on graphs: Extending high-dimensional data analysis to networks and other irregular domains,'' \emph{IEEE Signal Process. Mag.}, vol.~30, no.~3, pp. 83--98, 2013.

\bibitem{koppen2000curse}
M.~K{\"o}ppen, ``The curse of dimensionality,'' in \emph{5th Online World Conf. Soft Comput. Industr. Appl. (WSC5)}, vol.~1, 2000, pp. 4--8.

\bibitem{tomasi1998bilateral}
C.~Tomasi and R.~Manduchi, ``Bilateral filtering for gray and color images,'' in \emph{6th Int. Conf. Comput. Vis.}, 1998, pp. 839--846.

\bibitem{qi2017pointnetplusplus}
C.~R. Qi, L.~Yi, H.~Su, and L.~J. Guibas, ``PointNet++: Deep hierarchical feature learning on point sets in a metric space,'' in \emph{Adv. Neural Inf. Process. Syst. 30}, 2017, pp. 5099--5108.

\bibitem{guerrero2018pcpnet}
P.~Guerrero, Y.~Kleiman, M.~Ovsjanikov, and N.~J. Mitra, ``PCPNet learning local shape properties from raw point clouds,'' \emph{Comput. Graph. Forum}, vol.~37, no.~2, pp. 75--85, 2018.

\bibitem{9903481}
X.~Wang, X.~Fan, and D.~Zhao, ``PointFilterNet: A filtering network for point cloud denoising,'' \emph{IEEE Trans. Circuits Syst. Video Technol.}, vol.~33, no.~3, pp. 1276--1290, 2023.

\bibitem{10099466}
X.~Wang, W.~Cui, R.~Xiong, X.~Fan, and D.~Zhao, ``FCNet: Learning noise-free features for point cloud denoising,'' \emph{IEEE Trans. Circuits Syst. Video Technol.}, vol. 33, no. 11, pp. 6288--6301, 2023.

\bibitem{roveri2018pointpronets}
R.~Roveri, A.~C. {\"O}ztireli, I.~Pandele, and M.~Gross, ``PointProNets: Consolidation of point clouds with convolutional neural networks,'' \emph{Comput. Graph. Forum}, vol.~37, no.~2, pp. 87--99, 2018.

\bibitem{pistilli2020learningrobust}
F.~Pistilli, G.~Fracastoro, D.~Valsesia, and E.~Magli, ``Learning robust graph-convolutional representations for point cloud denoising,'' \emph{IEEE J. Sel. Topics Signal Process.}, vol.~15, no.~2, pp. 402--414, 2020.

\bibitem{kipf2016semi}
T.~N. Kipf and M.~Welling, ``Semi-supervised classification with graph convolutional networks,'' in \emph{5th Int. Conf. Learn. Rep.}, 2017.

\bibitem{chamberlain2021beltrami}
B.~Chamberlain, J.~Rowbottom, D.~Eynard, F.~Di~Giovanni, X.~Dong, and M.~Bronstein, ``Beltrami flow and neural diffusion on graphs,'' in \emph{Adv. Neural Inf. Process. Syst. 34}, pp. 1594--1609, 2021.

\bibitem{chen2018neuralode}
R.~T.~Q. Chen, Y.~Rubanova, J.~Bettencourt, and D.~Duvenaud, ``Neural ordinary differential equations,'' in \emph{Adv. Neural Inf. Process. Syst. 31}, pp. 6571--6583, 2018.

\bibitem{dupont2019augmented}
E.~Dupont, A.~Doucet, and Y.~W. Teh, ``Augmented neural ODEs,'' in \emph{Adv. Neural Inf. Process. Syst. 32}, 2019, pp. 3140--3150.

\bibitem{poli2019graph}
M.~Poli, S.~Massaroli, J.~Park, A.~Yamashita, H.~Asama, and J.~Park, ``Graph neural ordinary differential equations,'' \emph{arXiv preprint arXiv:1911.07532}, 2019.

\bibitem{weinan2017proposal}
E.~Weinan, ``A proposal on machine learning via dynamical systems,'' \emph{Commun. Math. Stat.}, vol.~1, no.~5, pp. 1--11, 2017.

\bibitem{lu2018beyond}
Y.~Lu, A.~Zhong, Q.~Li, and B.~Dong, ``Beyond finite layer neural networks: Bridging deep architectures and numerical differential equations,'' in \emph{Proc. 35th Int. Conf. Mach. Learn.}, 2018, pp. 3276--3285.

\bibitem{gasteiger2018predict}
J.~Gasteiger, A.~Bojchevski, and S.~G{\"u}nnemann, ``Predict then propagate: Graph neural networks meet personalized {PageRank},'' in \emph{7th Int. Conf. Learn. Rep.}, 2019.

\bibitem{zhu2021interpreting}
M.~Zhu, X.~Wang, C.~Shi, H.~Ji, and P.~Cui, ``Interpreting and unifying graph neural networks with an optimization framework,'' in \emph{Proc. Web Conf. 2021 (WWW`21)}, 2021, pp. 1215--1226.

\bibitem{xu2018representation}
K.~Xu, C.~Li, Y.~Tian, T.~Sonobe, K.-i. Kawarabayashi, and S.~Jegelka, ``Representation learning on graphs with jumping knowledge networks,'' in \emph{Proc. 35th Int. Conf. Mach. Learn.}, 2018, pp. 5453--5462.

\bibitem{liu2020towards}
M.~Liu, H.~Gao, and S.~Ji, ``Towards deeper graph neural networks,'' in \emph{Proc. 26th ACM SIGKDD Int. Conf. Knowl. Disc. Data Min.}, 2020, pp. 338--348.

\bibitem{defferrard2016convolutional}
M.~Defferrard, X.~Bresson, and P.~Vandergheynst, ``Convolutional neural networks on graphs with fast localized spectral filtering,'' in \emph{Adv. Neural Inf. Process. Syst. 29}, 2016, pp. 3844--3852.

\bibitem{bianchi2021graph}
F.~M. Bianchi, D.~Grattarola, L.~Livi, and C.~Alippi, ``Graph neural networks with convolutional {ARMA} filters,'' \emph{IEEE Trans. Pattern Anal. Mach. Intell.}, vol.~44, no.~7, pp. 3496--3507, 2021.

\bibitem{he2021bernnet}
M.~He, Z.~Wei, Z.~Huang, and H.~Xu, ``BernNet: Learning arbitrary graph spectral filters via Bernstein approximation,'' in \emph{Adv. Neural Inf. Process. Syst. 34}, 2021, pp. 14\,239--14\,251.

\bibitem{kimmel2000images}
R.~Kimmel, R.~Malladi, and N.~Sochen, ``Images as embedded maps and minimal surfaces: Movies, color, texture, and volumetric medical images,'' \emph{Int. J. Comput. Vis.}, vol.~39, no.~2, pp. 111--129, 2000.

\bibitem{lee2018introduction}
J.~M. Lee, \emph{Introduction to Riemannian Manifolds}, 2nd~ed.\hskip 1em plus 0.5em minus 0.4em\relax Springer Cham, 2018.

\bibitem{nomizu1961existence}
K.~Nomizu and H.~Ozeki, ``The existence of complete Riemannian metric,'' \emph{Proc. Am. Math. Soc.}, vol.~12, no.~6, pp. 889--891, 1961.

\bibitem{vaswani2017attention}
A.~Vaswani \emph{et~al.}, ``Attention is all you need,'' in \emph{Adv. Neural Inf. Process. Syst. 30}, 2017, pp. 5998--6008.

\bibitem{chamberlain2021grand}
B.~Chamberlain, J.~Rowbottom, M.~I. Gorinova, M.~Bronstein, S.~Webb, and E.~Rossi, ``Grand: Graph neural diffusion,'' in \emph{Proc. 38th Int. Conf. Mach. Learn.}, 2021, pp. 1407--1418.

\bibitem{xhonneux2020continuous}
L.-P. Xhonneux, M.~Qu, and J.~Tang, ``Continuous graph neural networks,'' in \emph{Proc. 37th Int. Conf. Mach. Learn.}, 2020, pp. 10\,432--10\,441.

\bibitem{zhao2019pairnorm}
L.~Zhao and L.~Akoglu, ``PairNorm: Tackling oversmoothing in gnns,'' in \emph{8th Int. Conf. Learn. Rep.}, 2020.

\bibitem{farouki2012bernstein}
R.~T. Farouki, ``The Bernstein polynomial basis: A centennial retrospective,'' \emph{Comput. Aided Geom. Des.}, vol.~29, no.~6, pp. 379--419, 2012.

\bibitem{he2010guided}
K.~He, J.~Sun, and X.~Tang, ``Guided image filtering,'' in \emph{11th Eur. Conf. Comput. Vis. (ECCV)}, 2010, pp. 1--14.

\bibitem{fan2017point}
H.~Fan, H.~Su, and L.~J. Guibas, ``A point set generation network for 3D object reconstruction from a single image,'' in \emph{2017 IEEE Conf. Comput. Vis. Pattern Recognit. (CVPR)}, 2017, pp. 605--613.

\bibitem{chang2015shapenet}
A.~X. Chang \emph{et~al.}, ``{ShapeNet}: An information-rich {3D} model repository,'' \emph{arXiv preprint arXiv:1512.03012}, 2015.

\bibitem{Visionair}
VisionAir, \url{http://www.infra-visionair.eu/} Accessed November 14, 2017.

\bibitem{gardner2003linear}
A.~Gardner, C.~Tchou, T.~Hawkins, and P.~Debevec, ``Linear light source reflectometry,'' \emph{ACM Trans. Graph.}, vol.~22, no.~3, pp. 749--758, 2003.

\bibitem{serna2014paris}
A.~Serna, B.~Marcotegui, F.~Goulette, and J.-E. Deschaud, ``Paris-rue-Madame database: a 3D mobile laser scanner dataset for benchmarking urban detection, segmentation and classification methods,'' in \emph{Proc. 3rd Int. Conf. Pattern Recognit. Appl. Methods}, 2014, pp. 819--824.

\bibitem{gschwandtner2011blensor}
M.~Gschwandtner, R.~Kwitt, A.~Uhl, and W.~Pree, ``BlenSor: Blender sensor simulation toolbox,'' in \emph{Int. Symp. Vis. Comput.}, 2011, pp. 199--208.

\bibitem{wang2019dynamic}
Y.~Wang, Y.~Sun, Z.~Liu, S.~E. Sarma, M.~M. Bronstein, and J.~M. Solomon, ``Dynamic graph CNN for learning on point clouds,'' \emph{ACM Trans. Graph.}, vol.~38, no.~5, article no. 146, pp. 1--12, 2019.

\bibitem{chen2022deep}
H.~Chen, S.~Luo, and W.~Hu, ``Deep point set resampling via gradient fields,'' \emph{IEEE Trans. Pattern Anal. Mach. Intell.}, vol.~45, no.~3, pp. 2913--2930, 2022.

\bibitem{wei2024pathnet}
Z.~Wei, H.~Chen, L.~Nan, J.~Wang, J.~Qin, and M.~Wei, ``PathNet: Path-selective point cloud denoising,'' \emph{IEEE Trans. Pattern Anal. Mach. Intell.}, vol. 46, no. 6, pp. 4426--4442, 2024.

\bibitem{wu2020comprehensive}
Z.~Wu, S.~Pan, F.~Chen, G.~Long, C.~Zhang, and S.~P. Yu, ``A comprehensive survey on graph neural networks,'' \emph{IEEE Trans. Neural Netw. Learn. Syst.}, vol. 32, no. 1, pp. 4--24, 2020.

\bibitem{li2018deeper}
Q.~Li, Z.~Han, and X.-M. Wu, ``Deeper insights into graph convolutional networks for semi-supervised learning,'' in \emph{Proc. 32nd AAAI Conf. Artif. Intell.}, 2018, pp. 3538--3545.

\bibitem{liao2019lanczosnet}
R.~Liao, Z.~Zhao, R.~Urtasun, and R.~S. Zemel, ``LanczosNet: Multi-scale deep graph convolutional networks,'' in \emph{7th Int. Conf. Learn. Rep.}, 2019.

\bibitem{zhou2005regularization}
D.~Zhou and B.~Sch\"{o}lkopf, ``Regularization on discrete spaces,'' in \emph{Proc. Joint Pattern Recognit. Symp.}, 2005, pp. 361--368.

\bibitem{diunderstanding}
F.~Di Giovanni, J.~Rowbottom, B.~P. Chamberlain, T.~Markovich, and M.~M. Bronstein, ``Understanding convolution on graphs via energies,'' \emph{Trans. Mach. Learn. Res.}, 2023.

\bibitem{cai2020note}  
C.~Cai and Y.~Wang, ``A note on over-smoothing for graph neural networks,'' 
\emph{arXiv preprint arXiv:2006.13318}, 2020.

\bibitem{RM291-297}
C.~B.~Sumathi and R.~Jothilakshmi, ``BIBO Stability and Decomposition Analysis of Signals and System with Convolution Techniques,'' \emph{Ratio Mathematica}, vol.~46, no.~0, 2023.

\end{thebibliography}
\end{document}